\documentclass{article}\usepackage[]{graphicx}\usepackage[]{xcolor}
\makeatletter
\def\maxwidth{ %
  \ifdim\Gin@nat@width>\linewidth
    \linewidth
  \else
    \Gin@nat@width
  \fi
}
\makeatother

\definecolor{fgcolor}{rgb}{0.345, 0.345, 0.345}
\newcommand{\hlkwb}[1]{\textcolor[rgb]{0.69,0.353,0.396}{#1}}%

\usepackage{framed}
\makeatletter
 {\par\unskip\endMakeFramed%
 \at@end@of@kframe}
\makeatother

\definecolor{shadecolor}{rgb}{.97, .97, .97}
\definecolor{messagecolor}{rgb}{0, 0, 0}
\definecolor{warningcolor}{rgb}{1, 0, 1}
\definecolor{errorcolor}{rgb}{1, 0, 0}

\usepackage{alltt}

\usepackage[letterpaper,top=2cm,bottom=2cm,left=3cm,right=3cm,marginparwidth=1.75cm]{geometry}

\usepackage{amsmath, amsfonts, amssymb}
\usepackage{graphicx}
\usepackage{bm}
\usepackage{dsfont}
\usepackage{hyperref}
\usepackage{booktabs}

\usepackage{float}

\usepackage[inline]{enumitem}

\usepackage{setspace}
\usepackage{parskip}
\usepackage{authblk}

\usepackage{soul}
\usepackage{xcolor}

\providecommand{\keywords}[1]
{
  \small	
  \textbf{\textit{Keywords:}} #1
}

\usepackage{biblatex} 
\title{Flusion: Integrating multiple data sources for accurate influenza predictions}
\author[1]{Evan L. Ray}
\author[1]{Yijin Wang}
\author[2]{Russell D. Wolfinger}
\author[1]{Nicholas G. Reich}
\affil[1]{Department of Biostatistics and Epidemiology, University of Massachusetts, Amherst, MA, United States}
\affil[2]{JMP Statistical Discovery, Cary, NC, United States}

\date{}
\IfFileExists{upquote.sty}{\usepackage{upquote}}{}
\begin{document}

\maketitle

\begin{abstract}
Over the last ten years, the US Centers for Disease Control and Prevention (CDC) has organized an annual influenza forecasting challenge with the motivation that accurate probabilistic forecasts could improve situational awareness and yield more effective public health actions.
Starting with the 2021/22 influenza season, the forecasting targets for this challenge have been based on hospital admissions reported in the CDC’s National Healthcare Safety Network (NHSN) surveillance system.
Reporting of influenza hospital admissions through NHSN began within the last few years, and as such only a limited amount of historical data are available for this target signal.
To produce forecasts in the presence of limited data for the target surveillance system, we augmented these data with two signals that have a longer historical record: 1) ILI+, which estimates the proportion of outpatient doctor visits where the patient has influenza; and 2) rates of laboratory-confirmed influenza hospitalizations at a selected set of healthcare facilities.
Our model, Flusion, is an ensemble model that combines two machine learning models using gradient boosting for quantile regression based on different feature sets with a Bayesian autoregressive model.
The gradient boosting models were trained on all three data signals, while the autoregressive model was trained on only data for the target surveillance signal, NHSN admissions; all three models were trained jointly on data for multiple locations.
In each week of the influenza season, these models produced quantiles of a predictive distribution of influenza hospital admissions in each state for the current week and the following three weeks; the ensemble prediction was computed by averaging these quantile predictions.
Flusion emerged as the top-performing model in the CDC's influenza prediction challenge for the 2023/24 season.
In this article we investigate the factors contributing to Flusion's success, and we find that its strong performance was primarily driven by the use of a gradient boosting model that was trained jointly on data from multiple surveillance signals and multiple locations.
These results indicate the value of sharing information across multiple locations and surveillance signals, especially when doing so adds to the pool of available training data.
\end{abstract}

\keywords{infectious disease, forecasting, gradient boosting, transfer learning}

\section{Introduction}
\label{sec:intro}

Since the early 2010s, short-term forecasting for infectious diseases has become an increasingly common activity, often through collaborations between governmental and industry or academic partners.
Starting during the 2013/2014 influenza season, the FluSight collaborative forecasting exercises organized by the US Centers for Disease Control and Prevention (CDC) have brought together teams of academic, industry and governmental researchers to produce probabilistic forecasts of influenza activity in the United States \cite{biggerstaff_results_2016}.
After a pause during the first two years of the COVID-19 pandemic, this forecasting exercise restarted in the spring of 2022 \cite{us_centers_for_disease_control_and_prevention_flusight_2023}.
FluSight typically involves over 20 different teams submitting forecasts each week, using different methodologies and sometimes different data sources.
Ensemble forecast techniques have been used by CDC and other groups to combine individual team submissions into a single consensus forecast, which typically has shown some of the most accurate performance overall \cite{mcgowan_collaborative_2019,reich_accuracy_2019}.
A primary motivation for the FluSight challenges is to carefully evaluate forecasts against real data and to use them to improve situational awareness and yield more effective public health actions \cite{lutz_applying_2019}.

Starting with FluSight seasons just following the COVID pandemic, new high-resolution and low-latency data streams that came online during the pandemic have been used as the ``ground truth'' target for forecasting exercises. 
Specifically, from 2022 through 2024, the forecasting targets for FluSight were based on hospital admissions reported in the CDC’s National Healthcare Safety Network (NHSN) surveillance system \cite{cdc_flusurveillance_overview}.
Because reporting of influenza hospital admissions through NHSN began during the COVID pandemic, the NHSN data have only a limited amount of historical information about influenza hospitalizations.
While the NHSN data surpassed previous surveillance systems in terms of providing a specific measure of influenza activity at fine spatial and temporal scales, the lack of an extensive history of data for model training introduced challenges to learning about seasonal patterns in influenza burden in the US.  
To address this challenge, for the 2023/2024 FluSight season we developed a new model, called \emph{Flusion}, which pulled in data from external data sources with a longer history of observations.  
This model was the top-performing model submitted to FluSight in the 2023/2024 season.  
The purpose of this paper is to describe the Flusion model and provide insights into what aspects of its design were associated with its strong performance.

Stimulated in large part by outbreak forecasting challenges organized by governmental agencies \cite[e.g., ][]{viboud_rapidd_2018,mcgowan_collaborative_2019,johansson_open_2019}, there has been substantial methodological development in the area of infectious disease forecasting over the last 10 years, both for stand-alone and ensemble modeling strategies \cite{yamana_superensemble_2016,reich_accuracy_2019,reis_superensemble_2019,ray_comparing_2023}.
Most successful stand-alone approaches can be characterized broadly as using statistical or machine learning methods in conjunction with conceptual models that are informed by infectious disease transmission dynamics, either explicitly through a compartmental transmission model or implicitly through inclusion of seasonality or observations from recent time-points \cite{reich_collaborative_2019,lopez2024covidCase}.
There has simultaneously been a large increase in worldwide participation in online data science competition sites like Kaggle, including many competitions with time series forecasting and even a series of five weeks of real time forecasting of COVID-19 in 2020 \cite{kaggle2020covidw1, kaggle2020covidw2, kaggle2020covidw3, kaggle2020covidw4, kaggle2020covidw5}.  The top performers in these competitions make heavy use of advanced feature engineering in conjunction with gradient boosted tree methods \cite{friedman2001gbm}, using the popular packages XGBoost, LightGBM, and CatBoost. For example, the majority of the top-performing competitors in the M5 forecast accuracy competition used gradient boosting \cite{makridakis2022m5}.

The Flusion model fits into the machine learning model category, although it uses features that emphasize both seasonality as well as recent observations and recent trends as a way of encoding underlying transmission dynamics of influenza.
In terms of underlying methodology, our primary focus in this work is on forecasting using gradient boosting. In particular, our work closely follows the approach of \cite{lainder2022forecastingGBT}, which was the leading approach in several time series forecasting competitions including strong performance for forecasting COVID-19 \cite{lopez2024covidCase}.

A key area where the Flusion model provides a methodological innovation is in how it leverages long historical time-series data that are closely related to the target data source that has little history.
There is existing work that focuses on situations where auxiliary data signals related to a primary signal of interest are used to improve short-term predictions.
For example, in settings where a primary signal has delayed or unreliable recent reporting, research has shown that other signals that are available contemporaneously or with a shorter lag can help inform estimation and prediction of trends \cite{yang_accurate_2015,farrow_modeling_2016,osthus_even_2019,leuba_tracking_2020,mcdonald_can_2021,jahja_real-time_2022}.
Some of these modeling efforts, among others, also highlight how sharing contemporaneous information across different locations can be helpful \cite{farrow_modeling_2016,mcdonald_can_2021,osthus_multiscale_2021}.

In contrast, the Flusion model, in addition to borrowing information across locations, augments a primary data signal with a very short history with other closely related signals that measure the same general system but for a longer history. 
This approach bears a close resemblance to multi-task learning and transfer learning, where models are trained on related tasks with a large number of available observations either simultaneously with training on more limited data for the primary task, or in a first stage that is followed by fine-tuning on data for the target task \cite{pan2009surveytransfer}.
In a forecasting context, these methods have been used in domains such as energy use and climate control \cite{Ribeiro2018transferbuildingenergy, Grubinger2017transferclimatecontrol}.
However, to our knowledge the literature contains only a few applications of this general approach to disease outbreak forecasting.  We are aware of two studies that investigated forecasting a pathogen of interest using data from other pathogens to inform model estimation \cite{Roster2022transfernewdiseases,Coelho2020transfermoquitoborne}. Zou et al.\ used multitask learning to forecast influenza-like illness by training models jointly on data for multiple geographic locations, and demonstrated that this was helpful in settings where some locations had only limited data but others had a longer history of observations \cite{zou2018multitaskILI}.
Unlike this existing literature, our work augments a signal measuring seasonal influenza in the US that has a limited history with other signals that measure the same disease system over a longer time span.
We note that contemporaneously with our development of Flusion, another group also explored methods for forecasting influenza hospitalizations as reported in NHSN using other data streams to extend the amount of available training data \cite{Meyer2024flutransfer}. Their approach differed from ours in that they used these other data streams to impute additional historical observations of hospitalizations, rather than training a model directly on data from multiple surveillance signals.
The success of the Flusion model and similar methods in the 2023/2024 FluSight season suggests that this approach may be a promising direction, especially in the context of public health data modernization initiatives that may replace long-standing surveillance data systems with new data streams.

The remainder of this article is structured as follows. We begin with some context for our work, including a discussion of the data sources we use in section \ref{sec:data} and an overview of the FluSight forecasting exercise in \ref{sec:flusight}. We establish notation in section \ref{sec:notation} and describe our modeling approaches in section \ref{sec:model}, and then we discuss the performance of our real-time forecasts in section \ref{sec:real-time-results}. Section \ref{sec:post-hoc-results} details a series of ablation studies that provide insight about which aspects of our modeling approach were important to its success. We conclude in section \ref{sec:discussion}.

\section{Data Sources}
\label{sec:data}

\begin{figure}[!ht]
    \centering
    \includegraphics[width=\textwidth]{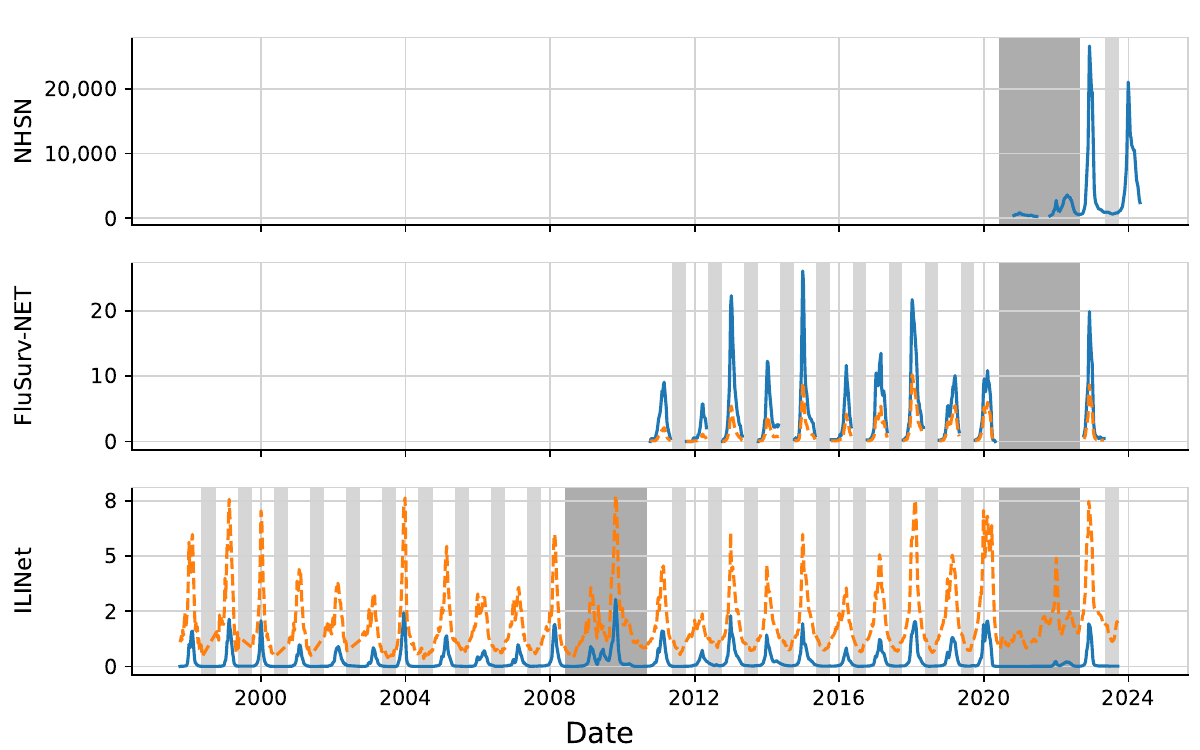}
    \caption{Influenza data at the national level in the US. The top panel shows weekly hospital admissions from NHSN, including the 2023/24 season that was the target season for predictions described in this article. In the second panel, a dashed orange line shows raw data reported from FluSurv-NET; the modeled data, in blue, are obtained by scaling up the raw data using per-season inflation factors designed to account for varying testing rates and test sensitivity.  In the third panel, a dashed orange line shows raw ILI data from ILINet; the modeled data in blue are ILI+ values obtained by combining ILI with test positivity rates. Dark grey shaded regions indicate pandemic seasons that were not used for model training; these include the 2008/09 and 2009/10 seasons which were impacted by pandemic H1N1 influenza, and the 2020/21 and 2021/22 seasons which were impacted by low influenza activity during the COVID pandemic. Light grey shaded regions indicate the off-season, which was not used for training the GBQR and GBQR-no-level models.  Additionally, FluSurv-NET and ILI+ data for the 2023/24 season were not used for model training in this work.}
    \label{fig:data_overview}
\end{figure}

Our model used three measures of influenza activity (Figure \ref{fig:data_overview}). The first of these was weekly influenza hospital admissions reported to the National Healthcare Safety Network (NHSN, \cite{cdc_flusurveillance_overview}), which was the target signal for FluSight (see section \ref{sec:flusight} for more detail). Reporting of influenza hospitalizations through NHSN began in 2020, and the first two years of reporting for this signal showed very low influenza activity during the COVID-19 pandemic. As a result, at the start of the 2023/24 season this signal had only one season's worth of data showing patterns typical of seasonal influenza.

To address this limitation of the target surveillance signal, our model used two other measures of influenza activity that have a longer reporting history. The first of these was a measure of hospital admissions where the patient has a positive influenza test as reported by FluSurv-NET, expressed as a rate per 100,000 population in the catchment areas of selected hospital facilities \cite{cdc_flusurvnet}.
The other signal was ILI+, which is an approximate measure of the proportion of outpatient doctor visits where the patient has influenza that is derived by combining data from ILINet and WHO/NREVSS \cite{goldstein2011-predicting-epidemic-sizes-flu-strains,cdc_flusurveillance_overview}.
As will be described further in section \ref{sec:model}, in this work our goal in using these data was to provide models with additional historical data to use for learning about trends in seasonal influenza activity. Because contemporaneous use of these signals was not our primary goal, and in order to avoid complications with integrating multiple data sources with different reporting lags and backfill behaviors, in this work we only used historical data prior to the 2023/24 season for these alternative data signals.

For both of these additional signals, we included adjustments designed to correct for known challenges with interpreting the data collected by FluSurv-NET and ILINet as consistent measures of influenza activity over time.  The FluSurv-NET data report on patients with a positive influenza test, and as such are subject to varying levels of underreporting depending on changing testing rates and test sensitivity from season to season.  The CDC produces annual estimates of hospital burden due to influenza that adjust for these factors \cite{cdc_flu_burden}.  We used these total burden estimates to estimate season-specific inflation factors that scale up reported rates from FluSurv-NET, with the intent of producing a more consistent measure of influenza activity over time; see supplemental section 2 for more details.  These inflation factors were generally larger in earlier seasons than in later seasons, indicating that FluSurv-NET undercounted influenza activity more in early seasons than it did in later seasons.

ILINet reports a measure of influenza-like illness (ILI), as the percent of outpatient doctor visits where the patient has symptoms consistent with influenza without another known cause. Because ILI is defined symptomatically, this signal generally includes some patients who have respiratory diseases other than influenza such as RSV and COVID. To address this, we computed ILI+ as the product of this ILI signal and influenza test positivity rates from laboratory testing sites reporting to World Health Organization (WHO) and National Respiratory Enteric Virus Surveillance System (NREVSS) systems.  After converting to a proportion scale, ILI+ can be interpreted as an estimate of the proportion of outpatient doctor visits where the patient has influenza, and has been used in previous forecasting work as a more specific measure of influenza activity than ILI \cite[e.g.][]{goldstein2011-predicting-epidemic-sizes-flu-strains,shaman2013-influenza-forecast-2012-2013}.

Our models were not trained on the 2008/09 and 2009/10 seasons, which were impacted by pandemic H1N1 influenza, and the 2020/21 and 2021/22 seasons, which were impacted by low influenza activity during the COVID pandemic. Additionally, the GBQR and GBQR-no-level models (see section \ref{subsec:model_gbqr}) were not trained on the influenza off-season. Designating US Epidemic week 31 as season week 1 (generally falling in early August), we define the off-season as season weeks less than 10 or greater than 40 \cite{nndss_2024}. Omission of the off-season from the training set forms a natural implementation of purging/embargoing, popular in econometric time series forecasting \cite{deprado2018financialML, lainder2022forecastingGBT}.

\section{The FluSight collaborative forecasting exercise}
\label{sec:flusight}

For the 2023/24 season, the primary target for forecasts collected in the FluSight forecasting exercise organized by CDC was weekly hospital admissions with confirmed influenza as reported in the NHSN data set for each of the 50 US states, the District of Columbia, Puerto Rico, and in total at the national level (Figure \ref{fig:data_overview}, top panel). Predictions for a second target representing a categorical measure of the direction and magnitude of change in admissions were also collected by FluSight, but we did not make predictions for that target with the Flusion model.

Predictions were submitted to FluSight on Wednesday each week. The week of submission is anchored relative to the Saturday after the submission date, which corresponds to the final day of that US epidemic week and is denoted as the \emph{reference date} for the predictions.
Predictions were made for hospital admissions from Sunday to Saturday in the current week and each of the three following weeks, corresponding to forecast horizons of 0, 1, 2, and 3 weeks ahead relative to the week of submission. Initially, FluSight also collected predictions at a horizon of -1, representing a ``hindcast" of admissions in the week before submission, but these hindcasts were discontinued a few weeks into the season and we do not analyze them here. A new data release from NHSN was made public at approximately noon on Wednesday each week including reported admissions up through the previous Saturday, and predictions were based on models fit to that data release. This data release was delayed by one day on the week ending on April 13, 2024, and for that week the forecast submission due date was extended so that the latest data could be used.

For the hospital admissions target, probabilistic predictions were represented with a set of predictive quantiles at $K = 23$ quantile levels $\alpha_k$, corresponding to a predictive median and the endpoints of 11 central prediction intervals at the nominal 10\%, ..., 90\%, 95\%, and 98\% levels. For example, the predictive median corresponds to $\alpha_{12} = 0.5$ and the 98\% interval corresponds to predictive quantiles at the levels $\alpha_1 = 0.01$ and $\alpha_{23} = 0.99$.

\section{Notation and evaluation metrics}
\label{sec:notation}

We use $z_{s, l, t}$ to denote the observed value of surveillance signal $s$ (with $s = 1$ for NHSN data, $s = 2$ for FluSurv data, and $s = 3$ for ILI+ data) in location $l$ in week $t$.  We denote the predictive quantile at level $\alpha_k$ for the value of the signal $s$ in location $l$, generated on reference date $d$ at forecast horizon $h$ by $q_{s,l,d,h,k}$.  Note that the observation $z_{s, l, t}$ at time $t$ is the prediction target across the four combinations of reference date $d$ and forecast horizon $h$ with $d + h = t$.  For brevity, we will often use the index $i$ to refer to a forecast task consisting of a combination of values of $s = s(i)$, $l = l(i)$, $d = d(i)$, and $h = h(i)$, with $q_{i,k}$ denoting the prediction at quantile level $\alpha_k$ for that task and $z_i$ denoting the corresponding observed value.

In this manuscript, we evaluate forecasts using three metrics: the mean absolute error (MAE) of the predictive median, the mean weighted interval score (MWIS), and the coverage rates of central 50\% and 95\% prediction intervals. In some figures, we will also examine one-sided quantile coverage rates. Our evaluations of forecast skill cover only predictions of NHSN admissions ($s = 1$).

MAE measures the average distance between the predictive median and the eventual observation, with smaller values indicating better forecast accuracy.  Recalling that the predictive median corresponds to the quantile level $\alpha_{12} = 0.5$ (i.e., $k = 12$), the absolute error of the prediction for task $i$ is $\vert q_{i, 12} - z_{i} \vert$.  In our evaluations, the mean absolute error averages the absolute error across predictions for different locations, dates, and horizons.

WIS can be viewed as a generalization of the absolute error to a set of quantile predictions, and is equivalent to an average of quantile scores (sometimes referred to as pinball losses) computed for each quantile prediction \cite{bracher2021WIS}:
\begin{align*}
WIS(\{q_{i,k}: k = 1, \ldots, K\}, z_{i}) &= \frac{1}{K} \sum_k 2 \cdot QS_{\alpha_k}(q_{i,k}, z_{i}) \\
QS_{\alpha_k}(q_{i,k}, z_{i}) &= \alpha_k \max(z_{i} - q_{i,k}, 0) + (1 - \alpha_k) \max(q_{i,k} - z_{i}, 0)
\end{align*}
The quantile score for a single quantile prediction assigns an asymmetric penalty to the distance between the prediction and the observation. The magnitudes of the penalties for underprediction and overprediction are set so that the expected value of the quantile score with respect to the forecast distribution is minimized by the quantile of the predictive distribution at the specified quantile level \cite{Gneiting2011quantilesoptimalpoint}. The quantile score for the quantile level 0.5 is equal to one half of the absolute error of the median. Again, lower values of WIS indicate a better alignment of forecasts with the observed data. MWIS is the average WIS across multiple locations, dates, and horizons.

We also compute relative versions of MAE and MWIS, denoted rMAE and rMWIS, using the ``pairwise tournament'' approach outlined in \cite{cramer2022covidMortalityForecasts}. The primary purpose of this procedure is to correct for the varying level of difficulty of the predictions submitted by different forecasters in settings where some forecasters did not provide predictions for all locations or time points. This is relevant to the evaluation of submissions to FluSight in section \ref{sec:real-time-results}, but in the experimental results in section \ref{sec:post-hoc-results} all forecasts were provided. A secondary goal is to standarize scores relative to a baseline model with known behavior, in our case the flat baseline described in section \ref{sec:real-time-results}. Smaller values of rMAE or rMWIS indicate better performance relative to the other models in the comparison pool, and in particular values less than 1 indicate performance that is better than the baseline.

In a comparison of forecast accuracy among $M$ models, computation of rMAE (rMWIS) has two steps. First, we obtain a summary of the average performance for model $m$ relative to each other model $m'$, denoted $\theta^m$. This summary is computed as the geometric mean of the ratio of the MAE (MWIS) for model $m$ to the MAE (MWIS) for each other model $m'$, where for each model pair the MAE averages across the set $\mathcal{I}_{m, m'}$ of locations, dates, and forecast horizons for which both models submitted predictions. The rMAE (or rMWIS) then normalizes this geometric mean relative to the value of $\theta^m$ for a baseline model:
\begin{align*}
rMAE^m &= \frac{\theta^m}{\theta^{baseline}},
\text{ where } \theta^m = \left( \prod_{m' \neq m} \frac{MAE_{\mathcal{I}_{m, m'}}^m}{MAE_{\mathcal{I}_{m, m'}}^{m'}} \right)^{1/(M-1)}
\end{align*}
Here, $MAE_{\mathcal{I}}^m$ denotes the MAE for model $m$ across all predictions for tasks $i$ in the index set $\mathcal{I}$.

Finally, we examine probabilistic calibration through the empirical coverage rates of 50\% and 95\% prediction intervals and individual quantiles. Let $l, u \in \{1, \ldots, K\}$ denote the indices for the lower and upper quantile levels defining the central prediction interval at a specified nominal level. For example, with our notation 95\% intervals correspond to $l = 2$ and $u = 22$, with $\alpha_l = 0.025$ and $\alpha_u = 0.975$. The empirical prediction interval coverage rate across prediction tasks in the index set $\mathcal{I}$ is defined as
$$\frac{1}{\vert \mathcal{I} \vert} \sum_{i \in \mathcal{I}} \mathds{1}(q_{i,l} \leq z_i \leq q_{i,u}),$$
where $\vert \mathcal{I} \vert$ is the size of the index set $\mathcal{I}$ and $\mathds{1}(q_{i,l} \leq z_i \leq q_{i,u})$ is $1$ if $q_{i,l} \leq z_i \leq q_{i,u}$ and $0$ otherwise.

One-sided coverage rates at individual quantile levels can provide more detail than interval coverage rates. In some figures below, we show the difference between the empirical one-sided coverage rate at each quantile level and the corresponding nominal coverage rate:
$$\delta_k = \frac{1}{\vert \mathcal{I} \vert} \sum_{i \in \mathcal{I}} \mathds{1}(z_i \leq q_{i,k}) - \alpha_k.$$
A well-calibrated forecaster will have $\delta_i \approx 0$, while a conservative forecaster with wide prediction intervals will have $\delta_k > 0$ for $\alpha_k > 0.5$ and $\delta_k < 0$ for $\alpha_k < 0.5$.

\section{Model}
\label{sec:model}

Flusion was constructed as an ensemble of statistical and machine learning time series models.  All models were fitted to data that were preprocessed to standardize across different signals, locations, and time points, and we describe these preprocessing steps in section \ref{subsec:data_standardization}.  We describe the component models in sections \ref{subsec:model_gbqr} and \ref{subsec:model_arx} and the ensemble methods in section \ref{subsec:model_ensemble}.  The precise formulation of the models included in the ensemble and the ensembling methods varied slightly over the course of the season, and we describe these aspects of our setup for generating real-time predictions in section \ref{subsec:model_realtime}.

\subsection{Data standarization}
\label{subsec:data_standardization}

\begin{figure}[!ht]
    \centering
    \includegraphics[width=\textwidth]{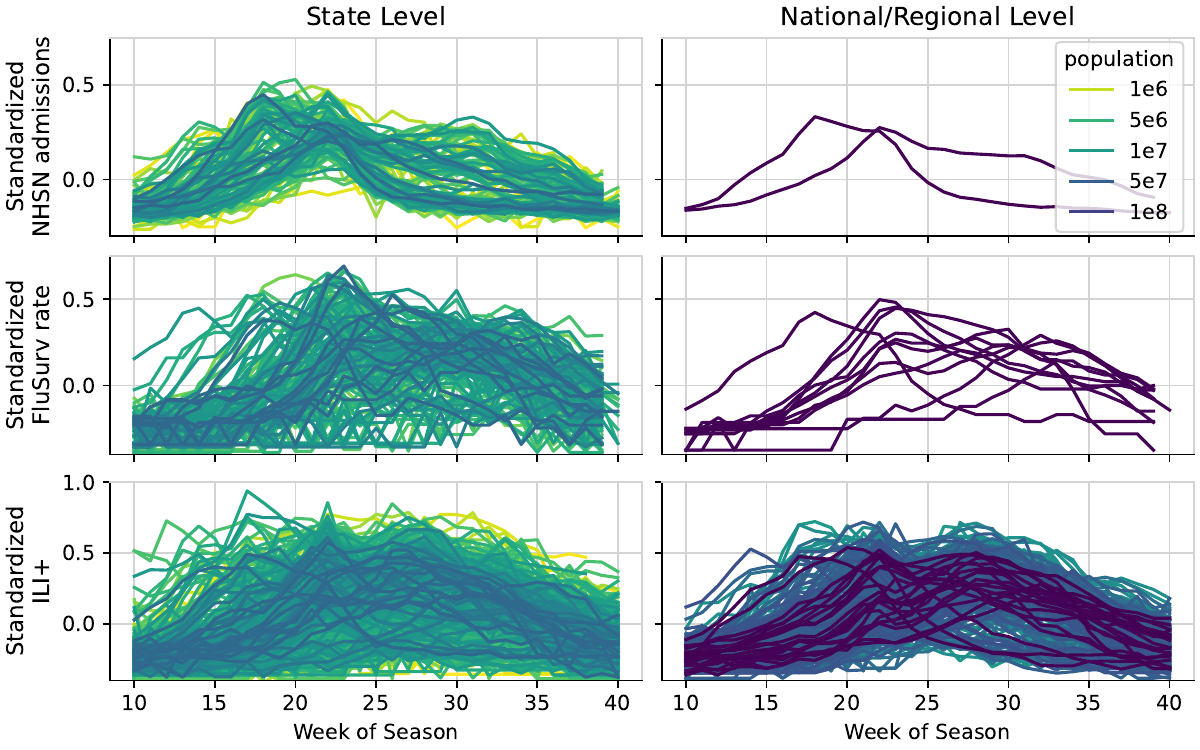}
    \caption{Influenza data for all surveillance signals and all locations available for each data source after standardizing transformations have been applied. The top row shows weekly hospital admissions from NHSN, the second row shows data from FluSurv-NET, and the third row shows ILI+. The left column shows all state-level locations, while the right column shows national level data, as well as data at the level of HHS regions for the ILI+ signal. The horizontal axis is the season week. We define the season to begin on US Epidemic Week 31, which generally falls in early August; the range of season weeks shown corresponds approximately to the active flu season. Within each panel, there is one line for each combination of season and location for all seasons and locations that are available for the given surveillance system at the state, regional, and national levels.  Line color corresponds to the population size of the location; the darkest lines are for the national level while the lightest lines are for states with small populations.}
    \label{fig:data_standardized}
\end{figure}

Our models were fitted to preprocessed versions of the surveillance data, with transformations designed to put the data on a similar scale for different surveillance signals and across different locations. Starting with the observed value $z_{s,l,t}$ of a particular surveillance signal $s$ in location $l$ at time $t$, we applied the following operations to compute the transformed version of the signal, $\tilde{z}_{s, l, t}$:
\begin{enumerate}
\item For NHSN admissions, we divided by the population of the location $l$ in units of 100,000 people to convert to a hospital admissions rate per 100,000 population, which is comparable across locations of different sizes. We used population counts reported by the US Census Bureau \cite{census_pop_older, census_pop_recent}. Note that the ILI+ and FluSurv signals are naturally expressed as rates or percents and so the magnitudes of those signals do not depend on population size.
\item We took a fourth root transformation to stabilize the variance of the signal across times of low and high influenza activity.
\item We scaled by dividing by the 95th percentile of all observations for each location and data source, and centered by subtracting the mean for each location and data source.  These transformations adjusted for varying magnitudes of the surveillance signals for different data sources and locations.
\end{enumerate}
The resulting transformed data used as an input to the models are shown in Figure \ref{fig:data_standardized}.

\subsection{Component Models 1 and 2: GBQR and GBQR-no-level}
\label{subsec:model_gbqr}

The first two models used gradient boosting for quantile regression, which we abbreviate as GBQR.
Specifically, we used GBQR to obtain separate quantile regression fits for each required quantile level $\alpha_k$, using the average quantile loss $QS_{\alpha_k}$ as the learning objective.
These models learned a mapping $f_{\alpha_k}(x_i)$ from features $x_i$ to the specified quantile of a predictive distribution for the target $y_i$, where $i$ indexes combinations of data source $s(i)$, location $l(i)$, reference date $d(i)$, and forecast horizon $h(i)$.  The estimated function $f_{\alpha_k}(x_i)$ takes the form of a sum of regression trees.  At each quantile level, the final prediction was obtained using bagging, by taking the median of predictive quantiles from 100 separate fits that were each based on a randomly selected 70\% of the seasons in the training set (including partial data for the current season).  The models were fitted using the LightGBM package in Python \cite{ke2017lightgbm} with default settings for all hyperparameters.

The models were trained jointly on data for all data sources, locations, dates, and forecast horizons.  However, the features $x_i$ contained information only about influenza activity for the particular data source $s(i)$ and location $l(i)$. Inclusion of multiple locations and data sources in the training data set allowed the model to use past examples from multiple locations and data sources to learn a mapping from $x$ to $y$. However, in our model setup, predictions of NHSN admissions in a particular location were not informed by contemporaneous observations of NHSN admissions in other locations or by contemporaneous observations of other surveillance signals in that same location. The use of contemporaneous observations from other locations or signals to inform predictions remains a topic for future work.

Both the features $x_i$ and the prediction targets $y_i$ were calculated based on the standardized version of the original surveillance signal, $\tilde{z}$ (see section \ref{subsec:data_standardization}).
As was described above, on reference date $d$ the most recent available data report on influenza activity in the previous week, $d-1$.
For forecast task $i$ with reference date $d(i)$ and forecast horizon $h(i)$, we defined the prediction target for this model to be $y_i := \tilde{z}_{l(i), s(i), d(i) + h(i)} - \tilde{z}_{l(i), s(i), d(i) - 1}$, the difference between the transformed signal value on the target date and on the date with most recent available reported data.
Thus, the model was trained to predict the change in influenza activity over the next $h(i) + 1$ time steps.
Predictions of this target were converted to the original scale by adding the last observed value ($\tilde{z}_{l(i), s(i), d(i) - 1}$) and inverting the initial data transformation operations described in section \ref{subsec:data_standardization}.

For the primary GBQR model, the feature vector $x_i$ contained 114 features as outlined in table \ref{tab:features}. These features included information about the data source and location being forecasted, the time of season when the forecast was generated, the forecast horizon, and measures of the local level, trend, and curvature of the surveillance signal in the weeks leading up to the time $d(i)$. Measures of local level of the signal included the signal value itself, rolling means of the signal in a trailing window, and the intercepts of Taylor polynomials fit to a trailing window of observations; these Taylor polynomial fits were also used to obtain the features measuring local trend and curvature. More information about the calculation of these features is given in supplemental section 3.

\begin{table}[htbp]
\centering
\begin{tabular}{cp{11cm}c}
\toprule
Group & \multicolumn{1}{l}{Description} & Count \\
\midrule
1 & A one-hot encoding of the data source. & 3 \\
\midrule
2 & A one-hot encoding of the location. & 65 \\
\midrule
3 & A one-hot encoding of the spatial scale of the location (``state", ``region", or ``national"). & 3 \\
\midrule
4 & The population of the location. & 1 \\
\midrule
5 & The week of the season with the most recent reported data, $d(i) - 1$. & 1 \\
\midrule
6 & The difference between the week of the season with the most recent reported data and Christmas week; for instance, a value of 3 means that the most recent data report is for the week three weeks after Christmas. & 1 \\
\midrule
7 & The forecast horizon. & 1 \\
\midrule
8 & The most recent reported value of the surveillance signal, for the time $d(i) - 1$. & 1 \\
\midrule
9 & The coefficients of a degree 2 Taylor polynomial fit to the trailing $w$ weeks of data, where $w \in \{4, 6\}$, with the reference point for the polynomial set to the time $d(i) - 1$.  These coefficients are estimates of the local level, first derivative, and second derivative of the signal at the time $d(i) - 1$. & 6 \\
\midrule
10 & The coefficients of a degree 1 Taylor polynomial fit to the trailing $w$ weeks of data, where $w \in \{3, 5\}$. These coefficients are estimates of the local level and first derivative of the signal at the time $d(i) - 1$. & 4 \\
\midrule
11 & The rolling mean of the signal over the last $w$ weeks, where $w \in \{2, 4\}$. & 2 \\
\midrule
12 & The values of all features from groups 8 through 11 at lags 1 and 2, representing estimates of the local level and first and second derivatives of the signal in each of the previous two weeks. & 26 \\
\bottomrule
\end{tabular}
\caption{Features used in the GBQR and GBQR-no-level models. There were 12 groups of features. The ``Count'' column gives the number of features in each group; there were 114 features in total across all groups. The GBQR-no-level model did not use the features from groups 8 through 12 that measured the local level of the signal. See supplemental section 3 for details of rolling mean and Taylor polynomial calculations.}
\label{tab:features}
\end{table}

Measures of the local level of the surveillance signal (i.e., the reported signal value, rolling means and the intercepts of Taylor polynomial fits) had a high feature importance in the primary GBQR model; see section \ref{sec:post-hoc-preprocessing} below and supplemental section 4 for more detail.  Starting in the eighth week of the season, on the reference date of December 2, 2023, we included a second variation on the GBQR model that was not allowed to see these ``local level" features. This was motivated by two considerations: (1) a model fit without features that had high importance in the primary GBQR model might introduce more model diversity to the Flusion ensemble; and (2) in seasons with particularly high or low incidence, measures of local level might not be a reliable indicator of the magnitude and direction of changes in future values of influenza activity. In experimental results below, we refer to this model variation as GBQR-no-level.

\subsection{Component Model 3: ARX}
\label{subsec:model_arx}

Our third component model was a Bayesian auto-regressive time series model with covariates (ARX). This model included only one covariate: a spike function indicating proximity to the week of Christmas, taking the value 3 on Christmas week, 2 in the week before and the week after Christmas, 1 two weeks before and two weeks after Christmas, and 0 otherwise. This covariate was intended to help the model account for the consistent peak in influenza activity that is observed near the holiday.

The ARX model was trained only on NHSN admissions, but as with the GBQR models, it was trained jointly on data for all locations. The autoregressive coefficients were shared across locations, while a separate variance parameter for the innovations was estimated for each location. Specifically, the model had the following structure, where we suppress the index $s$ and use $J$ to denote the autoregressive order (which we set to 8 in this work):
\begin{align*}
\tilde{Z}_{l,t} \mid \tilde{z}_{l, t - 1}, \ldots, \tilde{z}_{l, t - J}, x_{l, t - 1}, \ldots, x_{l, t - J}, \varepsilon_{l,t} &= \sum_{j = 1}^J \alpha_j \tilde{z}_{l, t - j} + \sum_{j = 1}^J \beta_j x_{l, t - j} + \varepsilon_{l, t} \\
X_{l,t} \mid x_{l, t - 1}, \ldots, x_{l, t - J}, \nu_{l, t} &= \sum_{j = 1}^J \gamma_j x_{l, t - j} + \nu_{l, t} \\
\varepsilon_{l, t} &\sim \text{Normal}(0, \sigma_{\varepsilon, l}) \\
\nu_{l, t} &\sim \text{Normal}(0, \sigma_{\nu, l})
\end{align*}
The parameters $\sigma_{\varepsilon, l}$ and $\sigma_{\nu, l}$, $l = 1, \ldots, L$, were assigned independent Half-Cauchy priors, and all of the $\alpha_j$, $\beta_j$, and $\gamma_j$ parameters were given the shared hierarchical normal prior $\alpha_j, \beta_j, \gamma_j \sim \text{Normal}(0, \xi)$ with a scale $\xi$ that followed a Half-Cauchy prior.

Note that unlike many ARX model specifications, our model did not take future values of the covariate $x_{l,t}$ as known, but rather it predicted the covariate alongside the primary prediction target. In our modeling setting, where $x_{l, t}$ was a deterministic function of the season week, this behavior was likely not ideal. Remedying this to allow for the provision of known future values of covariates is on a short list of model improvements to make. Anecdotally, we noted that near Christmas week the model's predictions of the covariate $x$ captured the spike function's behavior.

The motivation for sharing the auto-regressive coefficients across locations was that this might help prevent the model from overfitting to a limited amount of training data.  With an autoregressive order of $J=8$, there were a total of 24 $\alpha$, $\beta$, and $\gamma$ parameters to estimate (and even if the values of the covariate were held fixed, there would be 16 $\alpha$ and $\beta$ parameters to estimate).  Estimating these parameters based on only a single past season of data for one location would likely be infeasible.  This is similar to the strategy of joint estimation of shared parameters across multiple observed series that has been employed elsewhere \cite[e.g., ][]{Montero2021localglobalforecast}. On the other hand, we felt that it was important to estimate separate variance parameters for different locations because the amount of noise in the observed data varies substantially across locations with small and large populations (Figure \ref{fig:data_standardized}).

In this model, forecasts more than one step after the most recent observed data were obtained by iterating one-step-ahead predictions. As with the GBQR models, the model is specified in terms of the transformed data $\tilde{z}$, so predictions on the original scale of the data were obtained by inverting the initial data transformations described in section \ref{subsec:data_standardization}.

We used the the No-U-Turn Sampler (NUTS, \cite{hoffman2014nuts}) algorithm for model estimation in NumPyro \cite{phan2019composable}.

\subsection{Ensemble Methods: Quantile averaging}
\label{subsec:model_ensemble}

For prediction task $i$, each of the three models above produced predictive quantiles $q^m_{i,k}$, $k = 1, \ldots, 23$.  At each quantile level, the prediction for the Flusion model was the mean of the component predictive quantiles: $q_{i,k} = (1/M) \sum_m q^m_{i,k}$. This method has been referred to as quantile averaging \cite[e.g., ][]{lichtendahl2013betterAveProbQuant} or as Vincent ensembling after \cite{vincent1912functionsOfVibrissae}.

\subsection{Model adjustments used for real time forecasts}
\label{subsec:model_realtime}

Here we describe a few minor variations on the methodology outlined above that we introduced over the course of the season. The experimental results below indicate that these changes had only a minor impact on forecast performance.

As we described above, the GBQR and ARX component models were used throughout the full season, but the GBQR-no-level model was introduced starting in the eighth week. In the first week of the season, we used an additional model that obtained a predictive median using the same method as GBQR, but obtained predictions at other quantile levels by bootstrapping out-of-sample residuals. We discontinued use of this model from the second week on. Although we did not investigate formally, our anecdotal sense was that predictions from this model were too conservative (with wide prediction intervals), likely due to a strategy of sharing bootstrapped residuals across locations and surveillance signals with different signal-to-noise ratios.

In the first week of the season, we formed our submitted predictions by combining forecasts based on all available data and forecasts based on data up to the second-to-last observation (i.e., omitting the final reported value). This was because it was indicated that the latest available data were tentative and were subject to reporting corrections. In that instance, we used an equally-weighted linear pool (or distributional mixture) to combine the predictions based on the full data set and the partial data. From the second submission on, we submitted only the predictions based on all available data.

Finally, in the submission for reference date of December 16, 2023, we dropped our forecasts for Massachusetts. In that week, the most recent NHSN data release did not include full reporting for Massachusetts, and the FluSight organizers at CDC announced that they would not score or publicly communicate forecasts for Massachusetts that were submitted that week.

\subsection{Software availability and reproducibility}

Code for fitting models and generating predictions for real-time submissions and retrospective experiments is available at \url{https://github.com/reichlab/flusion}. Feature preprocessing functionality is implemented in a Python module at \url{https://github.com/reichlab/timeseriesutils/}, and the ARX model is implemented in a Python module at \url{https://github.com/elray1/sarix}. Code for analyses in the manuscript and supplement is available at \url{https://github.com/reichlab/flusion-manuscript}. The manuscript is generated with a reproducible workflow using Docker \cite{merkel2014docker} and knitr \cite{xie2014knitr}. Analyses were conducted using R version 4.4.0 \cite{RSoftware} and Python version 3.10.12 \cite{PythonSoftware}.

\section{Real-time performance: the 2023/24 FluSight season}
\label{sec:real-time-results}

In this section, we summarize model performance results for real-time submissions to FluSight in the 2023/24 season.

\subsection{Evaluation setup}

To avoid distortion of WIS and AE results, we did not evaluate forecasts that were made at the national level. Although FluSight originally allowed for collection of predictions at a horizon of -1 week, these were discontinued; our analysis includes predictions made at horizons of 0 weeks (nowcasts), and predictions at horizons of 1, 2, and 3 weeks ahead relative to the reference date.

We included all models that contributed forecasts for at least two thirds of the combinations of state-level locations, reference dates, and non-negative horizons for which FluSight collected forecasts over the course of the season. Because a comprehensive evaluation of all FluSight contributors is not the aim of this manuscript, we anonymized the names of other individually-contributed models in these results to focus attention on the comparisons that are of interest for our purposes.

FluSight produced two ensemble forecasts during the season: one using a quantile averaging approach and one using a linear pool. These two ensembles had very similar performance, though the linear pool had slightly better marginal calibration. However, the quantile averaging ensemble was used by CDC as the source of official communications throughout the season, and so we include results from only that ensemble here.

We also included results from two baseline methods. The first baseline, which we refer to as Baseline-flat, was a random walk model produced by FluSight (labeled as FluSight-baseline in submissions), which produced forecasts that extended from the most recent observation in a flat line, with expanding uncertainty based on historical differences in weekly hospital admissions. In this method, for each location $i$ the historical differences $\delta_{i,t} = y_{i,t} - y_{i,t-1}$ were collected, along with their negative values $-\delta_{i,t}$ (a process which we refer to as ``symmetrizing'' the differences). Forecasts at multiple step-ahead horizons were generated by iteratively sampling from this collection of symmetrized weekly differences.

The second baseline method, Baseline-trend, followed a similar process with a few modifications that were designed so that the resulting forecasts tended to follow the trend of recent observations. It was a quantile averaging ensemble of 16 variations on the baseline method. Most importantly, it incorporated variations that did not symmetrize the past differences, and rather than using all available history, it collected differences in a rolling window of the past few weeks. The 16 variations were obtained by using different options for the rolling window size, the temporal resolution of data used as an input (daily or weekly), a data transformation that was applied (no transformation or square root), and whether or not symmetrization was used. We emphasize that although this baseline was more methodologically involved than Baseline-flat, it produced an epidemiologically naive forecast that pushed forward a local estimate of the trend observed over the few most recent weeks. This model was named UMass-trends\_ensemble in real-time submissions to FluSight.

Scores were calculated based on the version of the NHSN target data signal that was released by FluSight on May 1, 2024.  Reporting of influenza admissions to NHSN was not mandatory beyond that date, and so this data release represents the latest available complete data release to use for forecast evaluation.

\subsection{Evaluation results}

\begin{figure}[!htb]
    \centering
    \includegraphics[width=\textwidth]{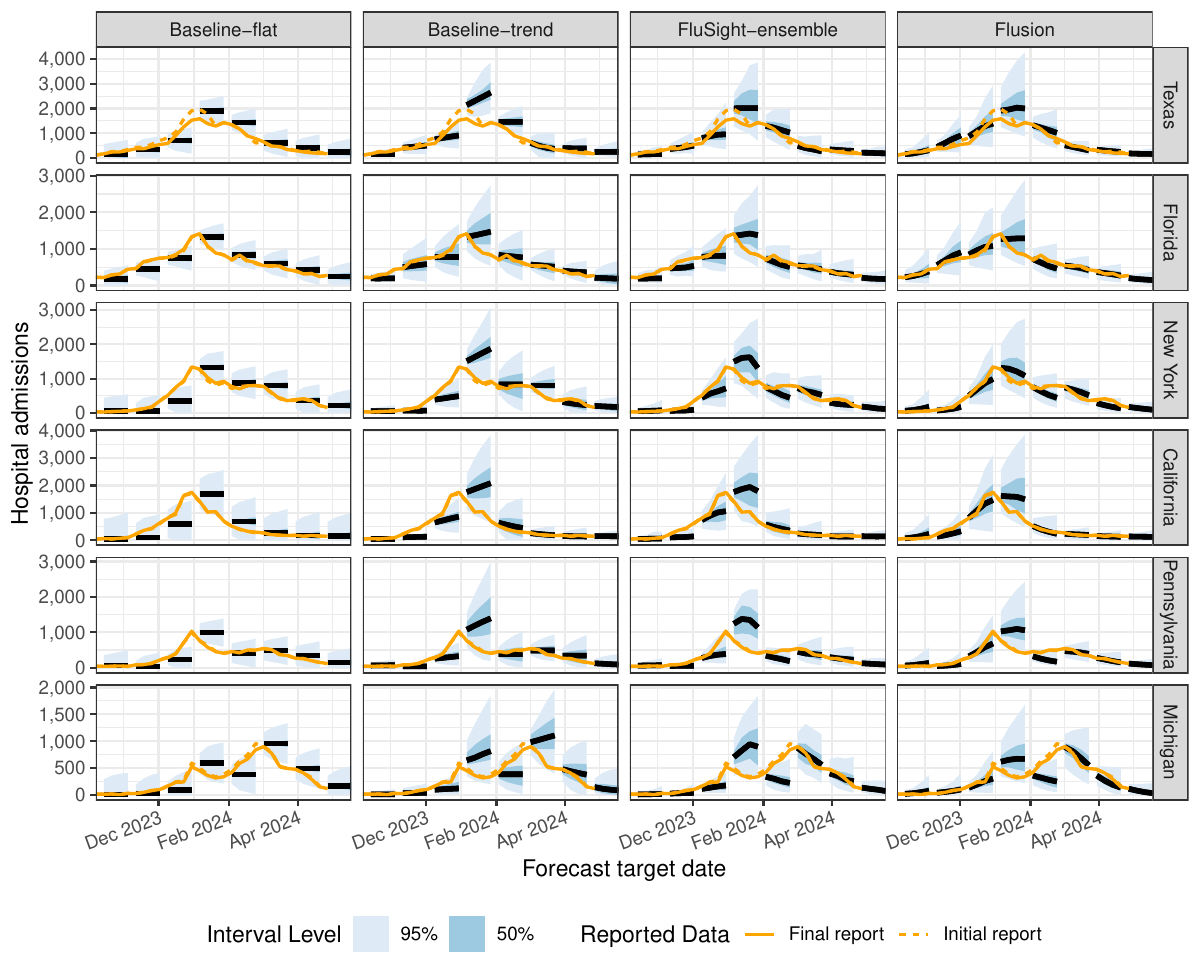}
    \caption{Influenza data and forecasts for the six states with the largest cumulative hospital admissions during the 2023/24 season. To avoid overplotting, in this figure forecasts from every fourth reference date are shown; evaluations include all reference dates. Forecasts are represented by the predictive median (black lines) and 50\% and 95\% prediction intervals (blue shaded regions). Solid orange lines show the finalized admission counts reported as of May 1, 2024, while dotted orange lines show the initial reported values that were available on the date predictions were generated.}
    \label{fig:forecasts_flusight}
\end{figure}

Forecasts from Flusion often appeared similar to forecasts from the FluSight-ensemble, though in several states (e.g. Florida, California, and New York) Flusion did a better job of capturing the increase in weekly hospital admissions in the early part of the season and a slightly better job of predicting the turnaround after the peak in late December (Figure \ref{fig:forecasts_flusight}). Qualitatively, both the FluSight-ensemble and Flusion generally captured trends in hospital admissions better than the baseline models during all phases of the season -- during the rise in the early part of the season, near the peak, and on the way down. An exception to this can be seen in forecasts for New York, Pennsylvania, and Michigan that were produced with a reference date of February 3, 2024. Those states had two local peaks: one near Christmas, and a second in late February or early March. This was a common pattern in many states in the northern US in the 2023/24 season, and in these instances Flusion (as well as the FluSight ensemble) typically produced incorrect predictions of continued decreases after the first peak heading into the second peak.

\begin{figure}[hpt]
    \centering
    \includegraphics[width=\textwidth]{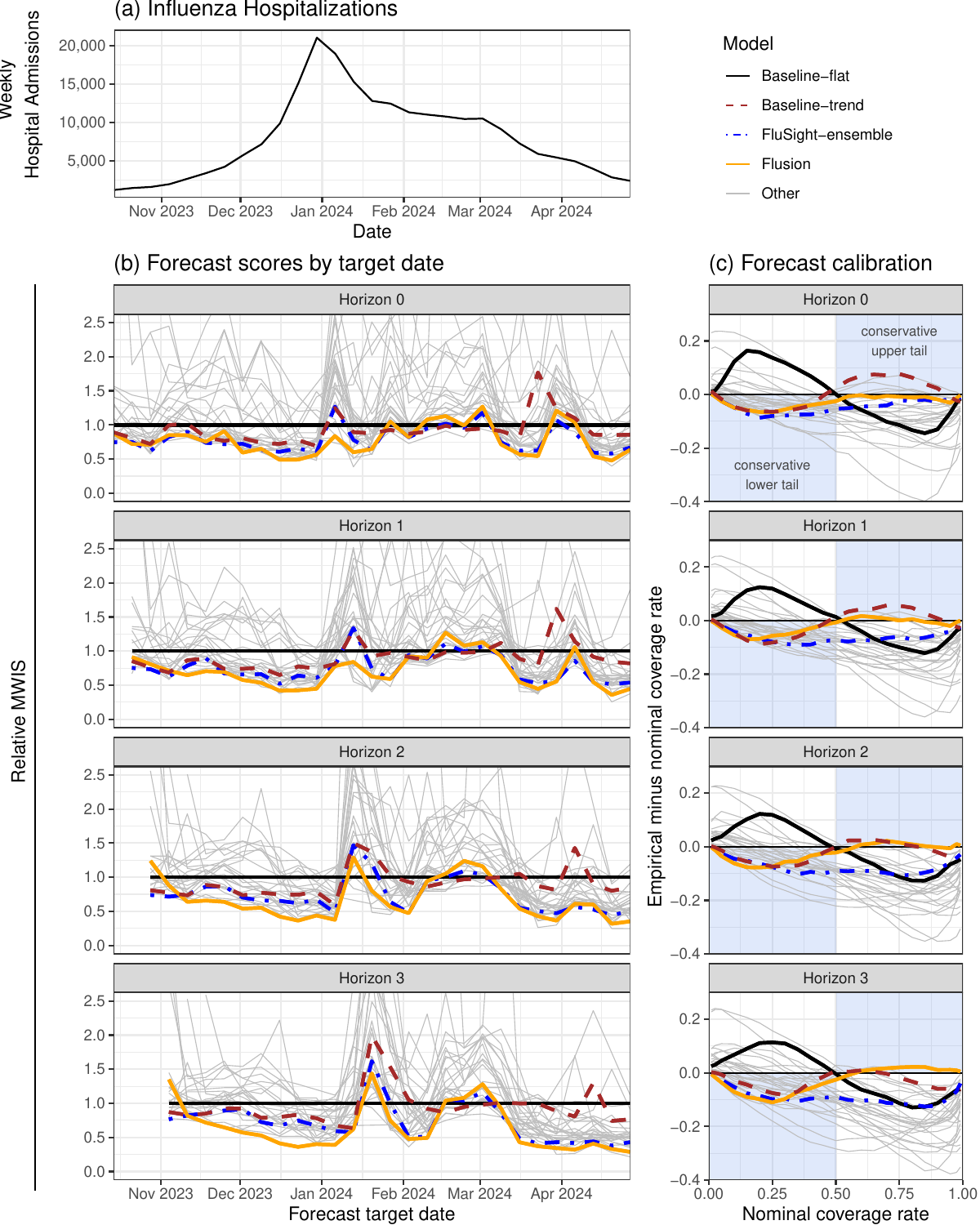}
    \caption{Influenza data and evaluation results. Panel (a): Weekly influenza hospital admissions reported in NHSN for the 2023/24 season, aggregated across all forecasted state-level locations. Panel (b): rMWIS for models contributing to FluSight, by forecast horizon (panels) and target date (horizontal axis). Lower rMWIS indicates better forecast performance. rMWIS values greater than 2.5 are not displayed. Panel (c): One-sided quantile coverage differential, computed as empirical coverage rate minus nominal coverage rate. A well-calibrated model has a differential of 0, while a conservative method (with wide prediction intervals) has a negative differential at nominal coverage rates less than 0.5 and a positive differential at nominal coverage rates greater than 0.5, indicated with blue shading.}
    \label{fig:scores_flusight}
\end{figure}

Aggregating across all forecast dates and forecast horizons, Flusion had the best performance as measured by rMWIS and rMAE among all models that contributed to FluSight for the 2023/24 season (Table \ref{tab:scores_flusight}). In a sensitivity analysis, we found that these results still held when the evaluation was conducted on the subset of forecasts generated for locations and reference dates for which the latest available data at the time of the forecast were not subsequently revised by 10 or more admissions (supplemental section 5). Flusion was consistently among the top-ranking models contributing to FluSight across individual forecast reference dates and forecast horizons (Figure \ref{fig:scores_flusight} (b)).

Prediction intervals from Flusion tended to be underconfident, i.e., prediction intervals were too wide on average (Table \ref{tab:scores_flusight}). An examination of one-sided quantile coverage rates indicates that marginally, the predictive quantiles in the upper tail of the forecast distribution are fairly well calibrated, while predictions for the lower quantile levels were too small on average (Figure \ref{fig:scores_flusight} (c)). Overall, the probabilistic calibration of Flusion was comparable to or better than that of other models contributed to FluSight, and it was superior to the calibration of the baseline and ensemble models.

\begin{table}[ht]
\centering
\begin{tabular}{lrrrrrrr}
  \toprule
Model & \% Submitted & MWIS & rMWIS & MAE & rMAE & 50\% Cov. & 95\% Cov. \\ 
  \midrule
\textbf{Flusion} & 99.9 & \textbf{29.6} & \textbf{0.610} & \textbf{45.6} & \textbf{0.670} & 0.583 & 0.967 \\ 
  \textbf{FluSight-ensemble} & 100.0 & 35.5 & 0.731 & 55.4 & 0.814 & 0.516 & 0.926 \\ 
  Other Model \#1 & 100.0 & 35.6 & 0.731 & 54.0 & 0.792 & 0.558 & \textbf{0.940} \\ 
  Other Model \#2 & 89.1 & 40.4 & 0.773 & 61.5 & 0.840 & 0.479 & 0.908 \\ 
  Other Model \#3 & 97.8 & 39.9 & 0.806 & 59.3 & 0.857 & 0.363 & 0.793 \\ 
  Other Model \#4 & 100.0 & 40.0 & 0.823 & 60.5 & 0.890 & \textbf{0.497} & 0.884 \\ 
  Other Model \#5 & 67.3 & 45.0 & 0.827 & 68.7 & 0.899 & 0.487 & 0.866 \\ 
  Other Model \#6 & 100.0 & 41.5 & 0.851 & 64.4 & 0.945 & 0.466 & 0.903 \\ 
  Other Model \#7 & 85.5 & 45.7 & 0.852 & 66.1 & 0.878 & 0.418 & 0.824 \\ 
  Other Model \#8 & 100.0 & 41.6 & 0.856 & 60.7 & 0.893 & 0.460 & 0.855 \\ 
  Other Model \#9 & 100.0 & 42.1 & 0.865 & 60.9 & 0.894 & 0.442 & 0.827 \\ 
  Other Model \#10 & 98.8 & 44.3 & 0.901 & 67.7 & 0.986 & 0.456 & 0.939 \\ 
  \textbf{Baseline-trend} & 99.9 & 43.9 & 0.906 & 67.0 & 0.990 & 0.618 & 0.922 \\ 
  Other Model \#11 & 95.7 & 45.0 & 0.908 & 66.2 & 0.956 & 0.554 & 0.870 \\ 
  Other Model \#12 & 87.0 & 45.0 & 0.936 & 70.7 & 1.050 & 0.449 & 0.929 \\ 
  Other Model \#13 & 96.4 & 42.4 & 0.948 & 64.2 & 1.030 & 0.429 & 0.896 \\ 
  Other Model \#14 & 93.6 & 48.7 & 0.980 & 70.8 & 1.020 & 0.473 & 0.838 \\ 
  Other Model \#15 & 99.2 & 47.3 & 0.993 & 58.1 & 0.870 & 0.596 & 0.793 \\ 
  \textbf{Baseline-flat} & 100.0 & 48.5 & 1.000 & 67.9 & 1.000 & 0.282 & 0.888 \\ 
  Other Model \#16 & 72.2 & 59.1 & 1.010 & 87.1 & 1.060 & 0.416 & 0.823 \\ 
  Other Model \#17 & 96.3 & 51.8 & 1.040 & 65.3 & 0.934 & 0.242 & 0.751 \\ 
  Other Model \#18 & 98.2 & 51.3 & 1.040 & 73.0 & 1.060 & 0.395 & 0.773 \\ 
  Other Model \#19 & 76.7 & 61.9 & 1.090 & 87.8 & 1.100 & 0.288 & 0.717 \\ 
  Other Model \#20 & 84.2 & 52.6 & 1.150 & 72.0 & 1.130 & 0.368 & 0.768 \\ 
  Other Model \#21 & 85.1 & 57.8 & 1.180 & 73.2 & 1.070 & 0.316 & 0.615 \\ 
  Other Model \#22 & 88.3 & 61.3 & 1.180 & 89.1 & 1.230 & 0.377 & 0.802 \\ 
  Other Model \#23 & 69.0 & 42.6 & 1.280 & 59.0 & 1.270 & 0.386 & 0.772 \\ 
  Other Model \#24 & 85.5 & 65.2 & 1.300 & 83.5 & 1.200 & 0.219 & 0.494 \\ 
  Other Model \#25 & 92.5 & 80.2 & 1.550 & 110.0 & 1.520 & 0.389 & 0.821 \\ 
  Other Model \#26 & 92.6 & 126.0 & 2.540 & 154.0 & 2.220 & 0.174 & 0.429 \\ 
   \bottomrule
\end{tabular}
\caption{Overall evaluation results for forecasts submitted to FluSight. Model names other than Flusion, FluSight-ensemble, Baseline-flat, and Baseline-trend are anonymized. The percent of all combinations of location, reference date, and horizon for which the given model submitted forecasts is shown in the ``\% Submitted" column; only models submitting at least 2/3 of forecasts were included. Results for the model with the best MWIS, rMWIS, MAE, and rMAE are highlighted. Results for the models where empirical PI coverage rates are closest to the nominal levels are highlighted.} 
\label{tab:scores_flusight}
\end{table}

\section{Post hoc model exploration}
\label{sec:post-hoc-results}

In this section, we investigate the degree to which the following aspects of Flusion contibuted to its strong performance:
\begin{enumerate*}
\item the formulation of Flusion as an ensemble of three individual models;
\item joint training on multiple data sets and multiple locations;
\item data preprocessing, including corrections for reporting inconsistencies in the ILINet and FluSurv-NET data, the use of a fourth root data transform, and the importance of the features that were used by the GBQR model.
\end{enumerate*}

\subsection{Component models and ensembling}

To investigate the skill of our individual component models and the added value of ensembling, we computed scores for each of the three component models that were members of the Flusion ensemble and for ensembles formed using two of the three components.  As documented in section \ref{sec:model}, the component models and ensembling method that we used in real time changed over the course of the season. To enable a clearer understanding of the contributions of these models, the results we present here are based on the specifications of the individual GBQR, GBQR-no-level, and ARX models and the quantile averaging ensemble method that were used for Flusion starting the week of December 2, 2023.  In instances where predictions from one or more component models were not created in real time, we created post hoc model fits and predictions using the data that would have been available in real time.

\begin{table}[htbp]
\centering
\begin{tabular}{lrrrrrrr}
\multicolumn{8}{l}{Experiment A: Component model performance} \\
  \toprule
Model & \% Submitted & MWIS & rMWIS & MAE & rMAE & 50\% Cov. & 95\% Cov. \\ 
  \midrule
GBQR, ARX & 100.0 & \textbf{29.9} & \textbf{0.618} & \textbf{45.3} & \textbf{0.668} & 0.570 & 0.958 \\ 
  Flusion & 100.0 & 30.2 & 0.622 & 46.6 & 0.686 & 0.558 & 0.963 \\ 
  GBQR & 100.0 & 30.3 & 0.625 & 46.3 & 0.682 & 0.529 & \textbf{0.947} \\ 
  GBQR, GBQR-no-level & 100.0 & 30.4 & 0.628 & 47.1 & 0.694 & 0.546 & 0.958 \\ 
  GBQR-no-level, ARX & 100.0 & 33.2 & 0.685 & 52.2 & 0.769 & 0.528 & 0.958 \\ 
  GBQR-no-level & 100.0 & 33.9 & 0.698 & 52.6 & 0.775 & 0.523 & 0.944 \\ 
  ARX & 100.0 & 39.5 & 0.815 & 60.0 & 0.884 & \textbf{0.485} & 0.917 \\ 
  Baseline-flat & 100.0 & 48.5 & 1.000 & 67.9 & 1.000 & 0.282 & 0.888 \\ 
   \bottomrule
\\
\multicolumn{8}{l}{Experiment B: Reduced training data} \\
  \toprule
Model & \% Submitted & MWIS & rMWIS & MAE & rMAE & 50\% Cov. & 95\% Cov. \\ 
  \midrule
Flusion & 100.0 & \textbf{30.2} & \textbf{0.622} & 46.6 & 0.686 & 0.558 & 0.963 \\ 
  GBQR & 100.0 & 30.3 & 0.625 & \textbf{46.3} & \textbf{0.682} & \textbf{0.529} & \textbf{0.947} \\ 
  GBQR-by-location & 100.0 & 37.8 & 0.780 & 57.9 & 0.854 & 0.327 & 0.891 \\ 
  GBQR-only-NHSN & 100.0 & 41.5 & 0.857 & 63.7 & 0.939 & 0.361 & 0.838 \\ 
  Baseline-flat & 100.0 & 48.5 & 1.000 & 67.9 & 1.000 & 0.282 & 0.888 \\ 
   \bottomrule
\\
\multicolumn{8}{l}{Experiment C: Data preprocessing} \\
  \toprule
Model & \% Submitted & MWIS & rMWIS & MAE & rMAE & 50\% Cov. & 95\% Cov. \\ 
  \midrule
GBQR-no-reporting-adj & 100.0 & \textbf{29.1} & \textbf{0.600} & \textbf{44.4} & \textbf{0.654} & 0.510 & 0.940 \\ 
  Flusion & 100.0 & 30.2 & 0.622 & 46.6 & 0.686 & 0.558 & 0.963 \\ 
  GBQR & 100.0 & 30.3 & 0.625 & 46.3 & 0.682 & 0.529 & 0.947 \\ 
  GBQR-no-transform & 100.0 & 31.1 & 0.642 & 48.0 & 0.708 & \textbf{0.497} & \textbf{0.948} \\ 
  Baseline-flat & 100.0 & 48.5 & 1.000 & 67.9 & 1.000 & 0.282 & 0.888 \\ 
   \bottomrule
\end{tabular}
\caption{Evaluation results for post hoc experiments investigating determinants of model performance.  Experiment A gives results for individual component models in the Flusion ensemble, ensembles of pairs of components, and the full Flusion ensemble including all three components.  Experiment B gives results for the GBQR model, which is trained jointly on data for all locations and data sources, and variations trained separately for each location (GBQR-by-location) and trained only on hospital admissions from NHSN (GBQR-only-NHSN).  Experiment C gives results for a variation on the GBQR model that does not incorporate reporting adjustments designed to improve the degree to which ILINet and FluSurv-NET data reflect influenza activity (GBQR-no-reporting-adj) and a variation that does not use a fourth-root transform (GBQR-no-transform), along with the original GBQR model which uses the reporting adjustments and the fourth-root transform. The percent of all combinations of location, reference date, and horizon for which the given model submitted forecasts is shown in the ``\% Submitted" column; in these retrospective experiments, we produced forecasts for all locations and time points. Within each experiment group, results for the model with the best MWIS, rMWIS, MAE, and rMAE are highlighted. Results for the models where empirical PI coverage rates are closest to the nominal levels are highlighted.} 
\label{tab:scores_experiments}
\end{table}

In this comparison, the most important determinant of performance was whether or not an ensemble included the GBQR model (Table \ref{tab:scores_experiments}, Experiment A). Score differences among the top four model variations were small, and all of those model variations included the GBQR model either alone or in combination with GBQR-no-level and/or ARX. There was a drop-off in performance for other variations that did not include GBQR. Thus, the fact that Flusion was constructed as an ensemble of methodologically distinct models was not a key driver of its performance. Indeed, GBQR alone and GBQR-no-level alone would each have placed first among all FluSight submissions if they had been submitted instead of Flusion, while ARX would have placed third among all contributing models. (We would note that gradient boosting methods are often considered to be ensemble methods, and remind the reader that we used bagging, another ensembling approach, within the GBQR method. Our observation here is that GBQR alone performed about as well as ensembles including GBQR alongside other distinct components.)

The GBQR-no-level model, which was trained without access to features measuring the local level of the time series, had worse performance than the primary GBQR model, and ensembles that included it generally performed slightly worse than ensembles that did not include it. We view this as evidence that the approach of omitting local level features was harmful to individual model performance without introducing enough differentiation from the primary GBQR model to serve as a useful ensemble member. In contrast, although ARX was the worst of our individual models, ensemble variations that included ARX were slightly better than ensemble variations that did not include it. Including models with more structural differences can be helpful in an ensemble, although in these results the gains in performance from including ARX were generally small.

\subsection{Joint training on multiple data sets and multiple locations}

The primary GBQR model was trained jointly on data from all three surveillance signals (NHSN, FluSurv-NET, and ILI+) and on data for all locations at the state, HHS regional, and US national level. To investigate the value of this joint model training approach, we considered two alternative methods:
\begin{enumerate}
\item The GBQR-only-NHSN model was trained on data for all locations, but using only hospital admissions from NHSN, the surveillance signal used as the prediction target.
\item The GBQR-by-location model was trained separately for each state-level jurisdiction using data for that location from all three data sources.
\end{enumerate}

Both of these alternatives underperformed relative to the GBQR model (Table \ref{tab:scores_experiments}, Experiment B). GBQR-by-location would have been among the top three contributing models to FluSight, and GBQR-only-NHSN would have been among the top ten contributing models; however, both would have underperformed relative to the FluSight ensemble. The decisions to train on multiple data sources and to train jointly on data for all locations were critical for achieving strong model performance.

\subsection{Data preprocessing}
\label{sec:post-hoc-preprocessing}

In a third experiment, we fitted two model variations to investigate the value of some of the data preprocessing steps we used.  The GBQR-no-reporting-adj model omitted the adjustments described in Section \ref{sec:data} that were intended to address reporting inconsistencies in the ILINet and FluSurv-NET data. Specifically, this model used the ILI signal directly rather than using test positivity rates to convert to ILI+, and it used the raw rates reported by FluSurv-NET rather than attempting to account for time-varying case capture rates in the FluSurv-NET system. In our evaluations, GBQR-no-reporting-adj outperformed the original GBQR model by a small amount (Table \ref{tab:scores_experiments}, Experiment C). In our model formulation, these reporting adjustments were not helpful to model performance and indeed the evidence suggests that they were counterproductive.

In a second model variation, we investigated whether or not the use of a fourth root data transform was helpful. The GBQR-no-transform model was fit to data without using a power transform, though other preprocessing steps described in section \ref{sec:data} were used, including converting hospital admissions to a rate per 100,000 population and applying centering and scaling operations to make the data more comparable across different locations and data sources. The GBQR-no-transform model had slightly worse performance than the original GBQR model, indicating that the power transform was helpful (Table \ref{tab:scores_experiments}, Experiment C).

We also investigated feature importance as measured by the number of times each feature was used for the splitting criterion in a tree node in the gradient boosting fits for the GBQR model (supplemental section 4). For this investigation, we used a representative fit from the reference date of January 6, 2024. We averaged the importance score across the gradient boosting fits from all 100 bags and all 23 quantile levels.  The top five features were the current season week, the population of the target location, the most recent observation of the surveillance signal (after preprocessing transformations), the forecast horizon, and the difference between the current season week and Christmas week.  These were followed by a group of features that also had fairly high importance, primarily consisting of features measuring the local level, trend, and curvature of disease incidence, as well as an indicator of whether the location was Puerto Rico (which sees substantively different trends in influenza activity than other locations), and indicators of what the data source was. A final group of features with lower importance included indicators for all other locations and indicators of the aggregation level for the location (state, regional, or national).

\section{Discussion}
\label{sec:discussion}

The Flusion model documented in this manuscript was the top-ranked model in the FluSight forecasting exercise for the 2023/24 season as measured by MAE and MWIS, and its probabilistic calibration was comparable to or better than that of other participating models. The experimental results presented here indicate that this strong performance was primarily driven by the use of a gradient boosting model that was trained jointly on data from multiple surveillance signals and locations. In contrast, other modeling decisions we made had a more minor impact on forecast accuracy. For example, forming predictions as an ensemble of the GBQR model and a more classical ARX model yielded only a small gain in performance, and our attempts to compensate for irregularities in reporting for FluSurv-NET and ILI data were counterproductive.

A limitation of the results presented in this manuscript is that they report on performance only for a single season in the United States. Ongoing evaluation of the methods we have outlined will be necessary to ensure that the strong performance we documented here generalizes across multiple influenza seasons. Additionally, it would be valuable to understand how our methods would perform in the face of an influenza pandemic. The strong performance of a similar model using gradient boosting to forecast COVID offers some reassurance on this front \cite{lopez2024covidCase}, but our reliance on a long history of training data representing seasonal influenza activity may impact model performance in other settings.

There are numerous avenues for improving on our methods, and we are pursuing some of these in future work. Our intuition is that it would be valuable to use contemporaneous observations of multiple signals, such as NHSN admissions, FluSurv-NET, and ILI together, to inform predictions of trends in disease incidence. Care will need to be taken with this since the relative magnitudes of these signals can vary across geographies and over time. Along these lines, it may also be possible to use other signals, such as insurance claims or internet activity, to improve forecast accuracy.

Although our models were trained jointly on data from multiple spatial units, the forecasts they produced were not directly informed by the spatial structure. We anticipate that improvements in accuracy could be achieved by including features that measure trends in flu activity in neighboring locations, or by reconciling predictions made at multiple hierarchical levels. These methods would likely be particularly useful for locations with small populations, which typically have a relatively low signal-to-noise ratio in reported surveillance data.

Our model could also be extended to take into account epidemiological understanding of disease transmission, such as measures of vaccine uptake and efficacy or the circulation of multiple strains of the influenza virus at different times over the course of the season.

We expect that the main insights presented in this work regarding the value of using data from multiple surveillance signals and locations could be useful in multiple modeling frameworks and for forecasting infectious diseases other than seasonal influenza. These results are of particular importance to the infectious disease forecasting community since they indicate a path forward in settings where new public health surveillance systems may come online and shut down in a span of a few years. In the absence of a long history of data for a signal of interest, borrowing information from similar data sets can provide important context for models to learn about patterns of disease transmission.

\section*{Acknowledgements}

We are grateful to Ryan Tibshirani and Logan Brooks for constructive comments on an initial version of this manuscript, and to Rebecca Sweger for code review.

This work has been supported by the National Institutes of General Medical Sciences (R35GM119582) and the U.S. CDC(1U01IP001122). The content is solely the responsibility of the authors and does not necessarily represent the official views of NIGMS, the National Institutes of Health, or CDC.

\printbibliography

\end{document}


\maketitle

\section{Introduction}

This document provides supplemental analyses and results for the Flusion manuscript. Section \ref{sec:flusurv_burden_adj} describes the approach we use to adjusting the FluSurv-NET data to account for testing rates and test sensitivity in that surveillance system.  Section \ref{sec:feats} describes the features used in the GBQR model that summarize local behavior of the target surveillance signal.  Section \ref{sec:feature_importance} presents results about the importance of the features used by the GBQR model. Sections \ref{sec:flusight_sensitivity} and \ref{sec:experiments_sensitivity} describe sensitivity analyses investigating whether the forecast evaluation results in the main text are impacted by data revisions.

\section{Reporting adjustments for FluSurv-NET data}
\label{sec:flusurv_burden_adj}

In this section, we describe the adjustments we made to the reported FluSurv-NET data. The measure of influenza activity reported by FluSurv-NET is the rate of positive influenza cases per 100,000 population in the catchment area of a reporting healthcare facility or group of facilities. For the purposes of FluSurv-NET, ``a case is defined as a person who is a resident in a defined FluSurv-NET catchment area and tests positive for influenza by a laboratory test ordered by a health care professional within 14 days prior to or during hospitalization'' \cite{cdc_flusurvnet}. This measure of influenza activity may be impacted by underdetection of influenza cases either if patients with influenza are not tested or if they are tested but the test generates a false negative result.

The US Centers for Disease Control and Prevention (CDC) produces annual estimates of influenza disease burden at the national level that adjust for testing rates and test sensitivity, including point estimates of total nationwide influenza hospitalizations in each season along with 95\% uncertainty intervals \cite{cdc_flu_burden_methods, cdc_flu_burden}. We used these to estimate a season-specific scale up factor $\alpha$ that was used to adjust the FluSurv-NET data. This factor was obtained by solving the following equation for $\alpha$ based on the total hospitalization rate over the course of the season that was reported across the entire FluSurv-NET network, the point estimate of national hospital burden due to influenza from CDC, and the US population in units of 100,000 people as reported by the US Census Bureau for the first year of the influenza season \cite{census_pop_older, census_pop_recent}:
$$\alpha \cdot (\text{cumulative reported hospitalization rate, FluSurv-NET})
= \frac{\text{National burden estimate}}{\text{100k US population}}$$

Table \ref{tab:flusurv_burden_adj} summarizes these terms and the resulting estimated scale-up factors for each season with FluSurv-NET data in our training set. Note that the scale-up factors are larger in earlier seasons than later seasons, indicating that data from FluSurv-NET undercounted influenza activity more in earlier seasons.

\begin{table}[ht]
\centering
\begin{tabular}{rrrrrr}
  \hline
Season & Cum. rate & US population & Est. burden (count) & Est. burden (rate) & $\alpha$ \\ 
  \hline
2010/11 & 21.7 & 309,321,666 &   290,000 & 93.8 & 4.3 \\ 
  2011/12 & 8.6 & 311,556,874 &   140,000 & 44.9 & 5.2 \\ 
  2012/13 & 44.0 & 313,830,990 &   570,000 & 181.6 & 4.1 \\ 
  2013/14 & 35.2 & 315,993,715 &   350,000 & 110.8 & 3.1 \\ 
  2014/15 & 64.0 & 318,301,008 &   590,000 & 185.4 & 2.9 \\ 
  2015/16 & 31.5 & 320,635,163 &   280,000 & 87.3 & 2.8 \\ 
  2016/17 & 62.0 & 322,941,311 &   500,000 & 154.8 & 2.5 \\ 
  2017/18 & 102.7 & 324,985,539 &   710,000 & 218.5 & 2.1 \\ 
  2018/19 & 63.5 & 326,687,501 &   380,000 & 116.3 & 1.8 \\ 
  2019/20 & 65.7 & 328,239,523 &   390,000 & 118.8 & 1.8 \\ 
  2022/23 & 62.4 & 333,287,557 &   475,000 & 142.5 & 2.3 \\ 
   \hline
\end{tabular}
\caption{Reported data, intermediate calculations, and final estimates for FluSurv-NET burden adjustments in each training season where we used FluSurv-NET data. The 'Cum. rate' column shows the cumulative reported hospitalization rate over the course of the season for the entire FluSurv-NET network. The US populaton column shows an estimate of the US population size from the US Census Bureau in the first year of the season (e.g., the value shown for the 2010/11 season is the population estimate for 2010). The 'Est. burden (count)' column shows the point estimate of influenza hospitalization burden produced by CDC for each season, and the `Est. burden (rate)` column expresses these burden estimates as a rate per 100,000 population in the US by dividing the estimated burden count by the US population in units of 100,000 people. The scale-up factor $\alpha$ is the ratio of the values in the 'Est. burden (rate)' and 'Cum. rate' columns.} 
\label{tab:flusurv_burden_adj}
\end{table}

\section{Features measuring local level, slope, and curvature of the surveillance signal}
\label{sec:feats}

As was described in section 5 of the main text, the GBQR models used features based on rolling means and the coefficients of Taylor polynomials fit to rolling windows of the data. These features are designed to estimate the local level, slope, and curvature of the surveillance signal at each point in time, and we describe their calculation here. Recall the notation $\tilde{z}_{l,s,t}$ representing the value of the signal for location $l$ and data source $s$ at time $t$, after some initial standardizing transformations as described in section 5.1 of the main text.

At time $t$, the rolling mean over the trailing window of length $w$ is computed as
\begin{equation}
\frac{1}{w} \sum_{u = t - w + 1}^t \tilde{z}_{l,s,u}. \label{eqn:roll_mean}
\end{equation}

The coefficients of a degree $d$ Taylor polynomial based on the trailing window of length $w$ relative to the anchor point $t$ are obtained by fitting the following model to the observations $\{\tilde{z}_{l,s,u}: u = t - w + 1, \ldots, t\}$:
\begin{align}
\tilde{z}_{l,s,u} &= \sum_{c = 0}^d \frac{1}{c!}\beta_c (u - t)^c + \varepsilon_u \label{eqn:taylor_model} \\
\varepsilon_u &\sim \text{Normal}(0, \sigma^2) \nonumber
\end{align}
For example, with $d = 2$ we fit the quadratic model
\begin{align*}
\tilde{z}_{l,s,u} &= \beta_0 + \beta_1 (u - t) + \frac{1}{2}\beta_2 (u - t)^2 + \varepsilon_u \\
\varepsilon_u &\sim \text{Normal}(0, \sigma^2)
\end{align*}
To motivate this, suppose that the underlying signal follows a mean trend over time given by the smooth function $g(u)$, with observation noise due to, e.g., the reporting process. The function $g$ can be written in terms of its derivatives $g^{(c)}$ using the Taylor expansion about the point $t$:
$$
g(u) = \sum_{c = 0}^\infty \frac{g^{(c)}(t)}{c!} (u - t)^c.
$$
Truncating to the first $d+1$ terms yields an approximation to $g$ in the neighborhood of $t$, and the coefficient estimates $\beta_c$ from the linear model \eqref{eqn:taylor_model} can be regarded as estimates of the corresponding derivatives $g^{(c)}(t)$.
We refer to estimates of $\beta_0$, $\beta_1$, and $\beta_2$ as estimates of the local level, trend, and curvature of the signal respectively. The highest degree we used in any of our feature computations was $d = 2$.
Note that the rolling mean of Equation \eqref{eqn:roll_mean} could also be obtained from this process using a Taylor polynomial of degree $d = 0$, though in practice we used a more direct implementation.

\begin{figure}[ht]
    \centering
    \includegraphics[width=\textwidth]{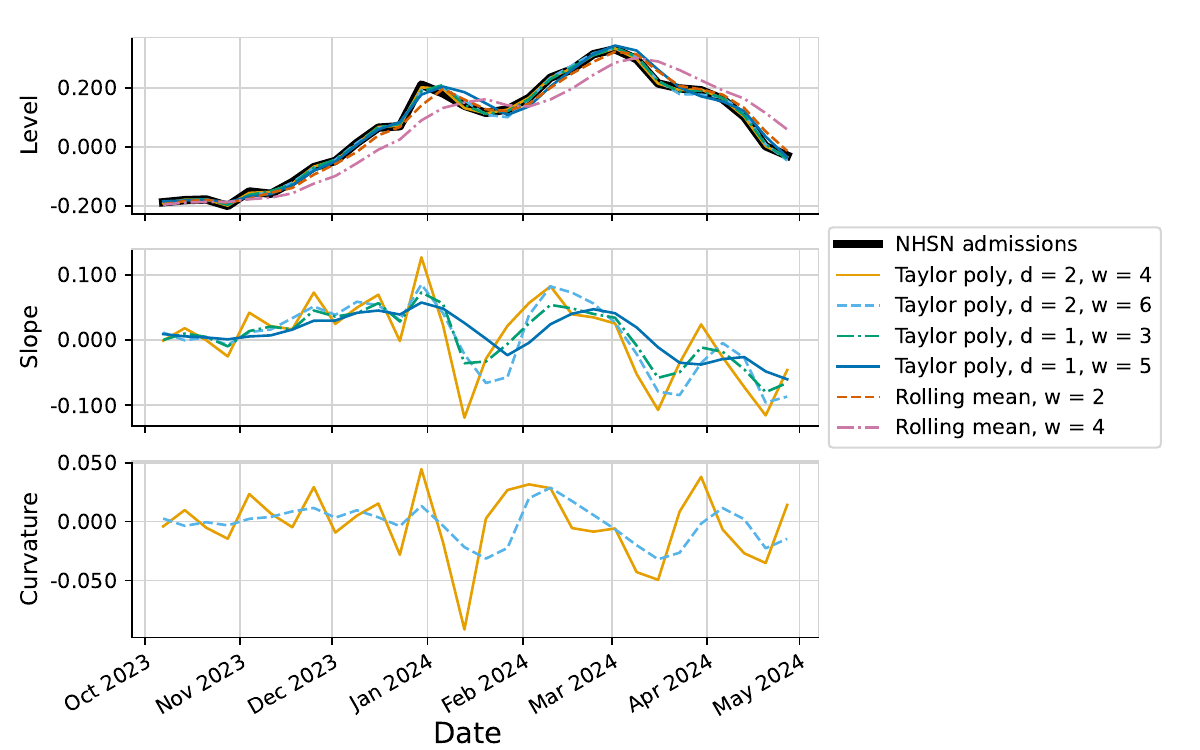}
    \caption{Example of features measuring the local level, trend, and curvature of the standardized NHSN admissions signal for the state of Michigan in the 2023/24 season (shown in black in the top panel for reference). At each time on the horizontal axis, a vertical line will intersect features calculated based on a trailing window ending on that date. For example, on Christmas week (just before Jan 2024), features based on a Taylor polynomial of degree $d=2$ fit to a trailing window of size $w = 4$ produced a local level estimate that closely matched the Christmas peak observed in the data, a positive slope just over 0.1 on the scale of the standardized data, and a positive curvature just under 0.05 indicating that the trend was increasing over that four week period.}
    \label{fig:features}
\end{figure}

Figure \ref{fig:features} illustrates the values of these features for the NHSN admission signal in the state of Michigan in the 2023/24 season. As expected, the features calculated based on longer window sizes $w$ and lower polynomial degrees $d$ vary more smoothly over time than features calculated based on shorter windows or higher polynomial degrees. Nevertheless, the features generally agree in terms of when the slope and curvature are positive or negative.

Note that at the end of the signal, only observations on or before the last time point are available. This motivates the use of a trailing window for feature calculation: with this choice, the features computed at both the end of the time series and at earlier time points can be expected to have similar characteristics as measures of local derivatives of the signal's trend. In contrast, if a centered window were used, estimates at earlier time points (when all observations within the centered window are available) would be more reliable than estimates at the end of the series.

Importantly, we do not account for the history of data revisions when we calculate these features. For example, for model fitting on reference date $t$, training examples are assembled for past times $u < t$ that include features measuring the local level, slope, and curvature at those times $u$. Those features are calculated based on the latest available data at time $t$, not based on the data that would have been available at time $u$. This means that our model implicitly estimates the relationships between these features and the target when the features are calculated on finalized, fully reported data. However, when predictions are generated extending from the reference date $t$, those features are calculated at the end of the time series when reported values more likely to be subsequently revised, leading to a mismatch between the data used for model fitting and the data used for prediction. This is a challenging problem to address in a setting like ours where the target data system has only a short reporting history and the characteristics of its revision process are not well known.

Finally, we highlight that although features such as the rolling mean or the intercept of a Taylor polynomial only directly measure the local level of the signal, when their lags are also included as features they can provide information about trend as well. For example, if we see that the rolling mean at time $t$ is larger than the rolling mean at time $t-1$ we may infer that the value of the signal is rising.

\section{Feature importance}
\label{sec:feature_importance}

\begin{figure}[htp]
    \centering
    \includegraphics[width=\textwidth]{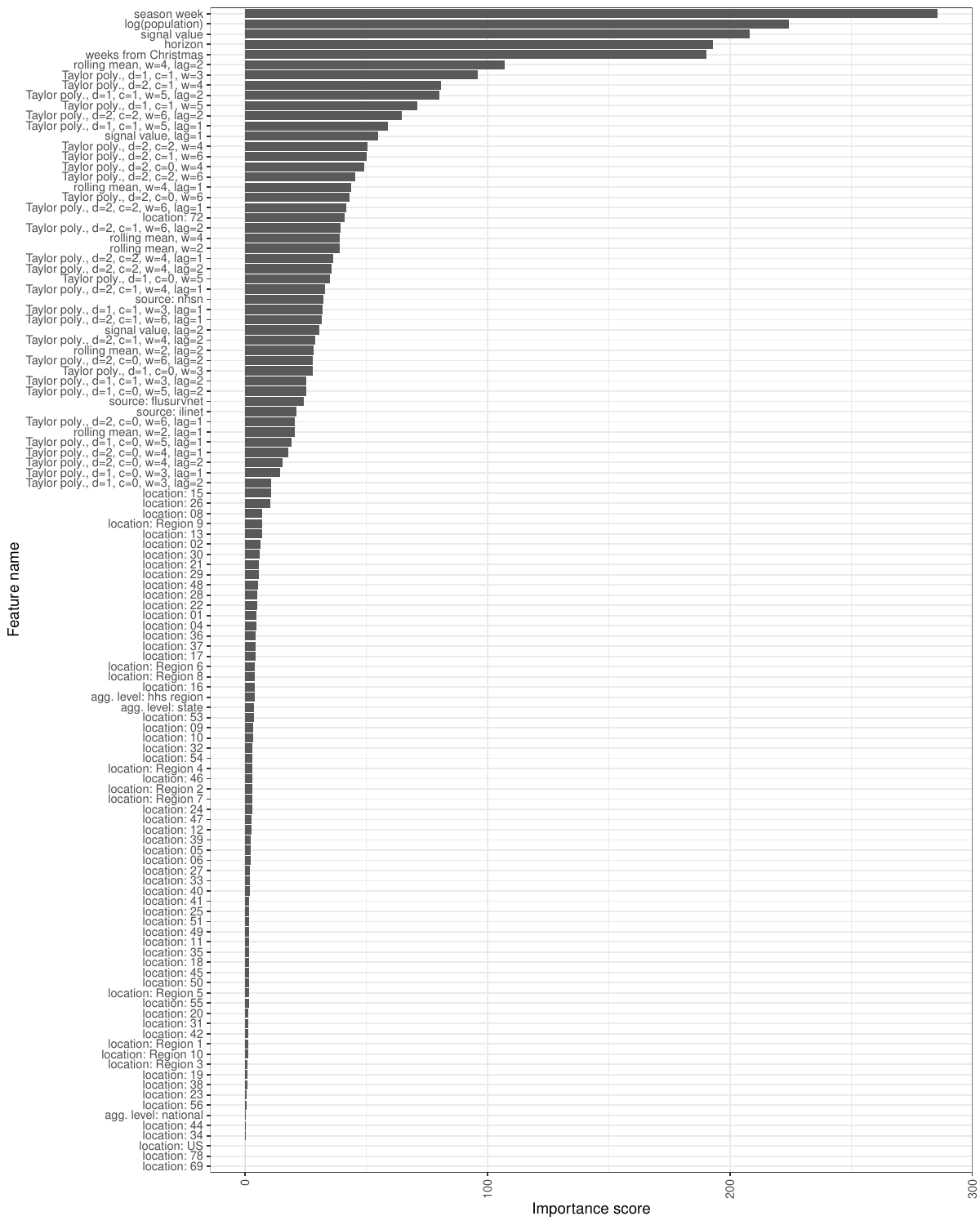}
    \caption{Feature importance values for the GBQR fit for the reference date of January 6, 2024. ``Signal value" indicates the latest reported value of the signal. For Taylor polynomial features, ``d'' indicates the polynomial degree and ``c'' indicates the coefficient corresponding to the feature value. For Taylor polynomial and rolling mean features, ``w'' indicates the window size used for feature calculation and if present ``lag" indicates the weekly lag used for the feature value.}
    \label{fig:feature_importance}
\end{figure}

Figure \ref{fig:feature_importance} shows feature importance scores for the GBQR fit for the reference date of January 6, 2024. See section 7 of the main text for more detail on how importance scores were calculated.

\section{FluSight results: sensitivity analysis for data revisions}
\label{sec:flusight_sensitivity}

Table \ref{tab:scores_flusight} contains MAE, MWIS, and PI coverage rates for real-time FluSight predictions, omitting predictions made on combinations of location and reference date for which the most recent available data at the time the prediction was generated were subsequently revised by 10 or more admissions.  This represents a generous sensitivity analysis, omitting 265 out of 1590 combinations of location and reference date for which predictions were submitted.  Figure \ref{fig:revisions-dropped-size-plot} displays information about the magnitudes of these revisions.

Comparing with Table 1 in the primary manuscript, we note that the main results discussed there still hold: Flusion has the best MAE and MWIS values by a substantial margin, while the marginal coverage rates of its central prediction intervals are too conservative.

\begin{table}[ht]
\centering
\begin{tabular}{lrrrrrrr}
  \hline
Model & \% Submitted & MWIS & RWIS & MAE & RAE & 50\% Cov. & 95\% Cov. \\ 
  \hline
\textbf{Flusion} & 100.0 & \textbf{21.1} & \textbf{0.575} & \textbf{32.3} & \textbf{0.626} & 0.597 & 0.971 \\ 
  Other Model \#1 & 100.0 & 26.2 & 0.714 & 39.3 & 0.762 & 0.565 & 0.939 \\ 
  \textbf{FluSight-ensemble} & 100.0 & 26.1 & 0.715 & 40.3 & 0.784 & 0.521 & 0.930 \\ 
  Other Model \#2 & 89.3 & 28.8 & 0.732 & 42.9 & 0.773 & 0.495 & 0.912 \\ 
  Other Model \#3 & 97.5 & 28.4 & 0.760 & 42.4 & 0.809 & 0.364 & 0.795 \\ 
  Other Model \#4 & 100.0 & 28.3 & 0.773 & 43.3 & 0.844 & \textbf{0.498} & 0.882 \\ 
  Other Model \#5 & 100.0 & 29.1 & 0.796 & 43.1 & 0.837 & 0.491 & 0.881 \\ 
  Other Model \#6 & 85.1 & 32.8 & 0.811 & 47.1 & 0.826 & 0.421 & 0.827 \\ 
  Other Model \#7 & 100.0 & 30.3 & 0.828 & 47.0 & 0.913 & 0.474 & 0.902 \\ 
  Other Model \#8 & 100.0 & 30.3 & 0.829 & 43.9 & 0.854 & 0.447 & 0.830 \\ 
  Other Model \#9 & 98.7 & 31.8 & 0.855 & 48.1 & 0.924 & 0.473 & \textbf{0.944} \\ 
  Other Model \#10 & 95.3 & 33.1 & 0.885 & 48.7 & 0.927 & 0.575 & 0.881 \\ 
  \textbf{Baseline-trend} & 100.0 & 32.5 & 0.890 & 49.5 & 0.967 & 0.639 & 0.929 \\ 
  Other Model \#11 & 87.3 & 32.1 & 0.902 & 49.5 & 0.988 & 0.443 & 0.923 \\ 
  Other Model \#12 & 96.9 & 31.7 & 0.925 & 47.3 & 0.984 & 0.429 & 0.892 \\ 
  Other Model \#13 & 92.8 & 34.7 & 0.937 & 49.8 & 0.955 & 0.464 & 0.829 \\ 
  Other Model \#14 & 95.9 & 35.7 & 0.945 & 46.8 & 0.881 & 0.242 & 0.778 \\ 
  Other Model \#15 & 98.1 & 36.3 & 0.976 & 51.2 & 0.981 & 0.393 & 0.772 \\ 
  Other Model \#16 & 68.3 & 43.1 & 0.982 & 63.7 & 1.030 & 0.416 & 0.829 \\ 
  Other Model \#17 & 99.2 & 35.2 & 0.982 & 42.9 & 0.850 & 0.580 & 0.789 \\ 
  \textbf{Baseline-flat} & 100.0 & 36.4 & 1.000 & 51.2 & 1.000 & 0.308 & 0.903 \\ 
  Other Model \#18 & 74.0 & 43.4 & 1.020 & 62.1 & 1.030 & 0.304 & 0.739 \\ 
  Other Model \#19 & 88.5 & 41.5 & 1.070 & 60.7 & 1.110 & 0.383 & 0.814 \\ 
  Other Model \#20 & 85.5 & 42.5 & 1.150 & 54.1 & 1.050 & 0.327 & 0.632 \\ 
  Other Model \#21 & 85.3 & 40.9 & 1.200 & 54.8 & 1.150 & 0.379 & 0.770 \\ 
  Other Model \#22 & 72.4 & 33.8 & 1.320 & 46.5 & 1.300 & 0.398 & 0.783 \\ 
  Other Model \#23 & 85.8 & 50.8 & 1.350 & 64.8 & 1.230 & 0.226 & 0.508 \\ 
  Other Model \#24 & 91.5 & 52.6 & 1.360 & 72.1 & 1.320 & 0.404 & 0.825 \\ 
  Other Model \#25 & 92.5 & 89.5 & 2.390 & 110.0 & 2.080 & 0.184 & 0.443 \\ 
   \hline
\end{tabular}
\caption{Overall evaluation results for forecasts submitted to the FluSight Forecast Hub, omitting forecasts made on combinations of reference date and location for which the latest available NHSN data at the time of the forecast were subsequently revised by 10 or more admissions. Model names other than Flusion, FluSight-ensemble, Baseline-flat, and Baseline-trend are anonymized. The percent of all combinations of location, reference date, and horizon for which the given model submitted forecasts is shown in the ``\% Submitted" column; only models submitting at least 2/3 of forecasts were included. Results for the model with the best MWIS, RWIS, MAE, and RAE are highlighted. Results for the models where empirical PI coverage rates are closest to the nominal levels are highlighted.} 
\label{tab:scores_flusight}
\end{table}

\begin{knitrout}
\definecolor{shadecolor}{rgb}{0.969, 0.969, 0.969}\color{fgcolor}\begin{figure}
\includegraphics[width=\maxwidth]{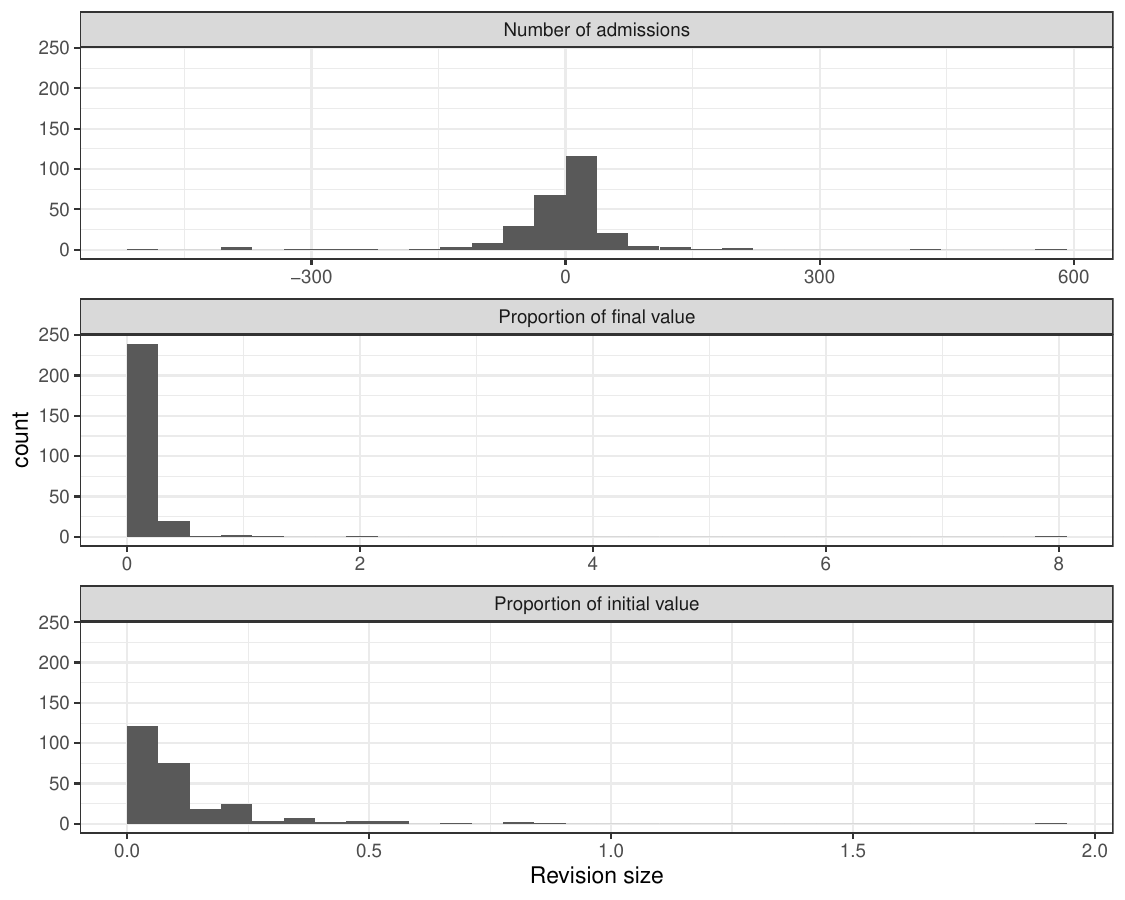} \caption[Measures of the size of reporting revisions for combinations of location and reference date that were omitted in the sensitivity analysis]{Measures of the size of reporting revisions for combinations of location and reference date that were omitted in the sensitivity analysis.  For legibility, only those revisions that were dropped (i.e., where the revision amount was at least 10 admissions up or down from the initial reported value) are displayed; most revisions were small.  The top panel shows the size of the revision in units of hospital admissions, where positive numbers indicate an upward revision of the initially reported value.  The second panel shows the absolute value of the revision size as a proportion of the final reported value.  The third panel shows the absolute value of the revision size as a proportion of the initial reported value.  When computing proportions, we add one to the denominator to avoid division by zero.  As an example, for October 7, 2023 (the last date for which data were available when producing predictions with a reference date of October 14, 2023), in Washington state the initial reported value was 43, which was subsequently revised down to a final value of 4. The revision amount is -39, which is 7.80 when expressed as a proportion of the final reported value or 0.89 when expressed as a proportion of the initial reported value.}\label{fig:revisions-dropped-size-plot}
\end{figure}

\end{knitrout}

\section{Experimental results: sensitivity analysis for data revisions}
\label{sec:experiments_sensitivity}

Table \ref{tab:scores_experiments} contains results from the post hoc experiments described in section 7 of the main text, omitting forecasts produced for combinations of location and reference date where the latest available NHSN data as of the reference date were subsequently revised up or down by at least 10 admissions. Comparing with table 2 of the main text, we see that the qualitative modeling results discussed there still hold in this sensitivity analysis.

\begin{table}[ht]
\centering
\begin{tabular}{lrrrrrrr}
\multicolumn{8}{l}{Experiment A: Component model performance} \\
  \hline
Model & \% Submitted & MWIS & RWIS & MAE & RAE & 50\% Cov. & 95\% Cov. \\ 
  \hline
GBQR, ARX & 100.0 & \textbf{21.2} & \textbf{0.582} & \textbf{32.0} & \textbf{0.626} & 0.589 & 0.967 \\ 
  GBQR & 100.0 & 21.3 & 0.583 & 32.4 & 0.632 & 0.544 & 0.952 \\ 
  Flusion & 100.0 & 21.4 & 0.588 & 32.9 & 0.642 & 0.574 & 0.969 \\ 
  GBQR, GBQR-no-level & 100.0 & 21.5 & 0.589 & 32.9 & 0.643 & 0.555 & 0.961 \\ 
  GBQR-no-level, ARX & 100.0 & 23.7 & 0.651 & 37.0 & 0.723 & 0.542 & 0.964 \\ 
  GBQR-no-level & 100.0 & 24.0 & 0.657 & 36.7 & 0.717 & 0.525 & \textbf{0.949} \\ 
  ARX & 100.0 & 28.1 & 0.771 & 43.0 & 0.841 & \textbf{0.508} & 0.934 \\ 
  Baseline-flat & 100.0 & 36.4 & 1.000 & 51.2 & 1.000 & 0.308 & 0.903 \\ 
   \hline
\\
\multicolumn{8}{l}{Experiment B: Reduced training data} \\
  \hline
Model & \% Submitted & MWIS & RWIS & MAE & RAE & 50\% Cov. & 95\% Cov. \\ 
  \hline
GBQR & 100.0 & \textbf{21.3} & \textbf{0.583} & \textbf{32.4} & \textbf{0.632} & \textbf{0.544} & \textbf{0.952} \\ 
  GBQR-by-location & 100.0 & 25.5 & 0.701 & 39.3 & 0.768 & 0.340 & 0.895 \\ 
  GBQR-only-NHSN & 100.0 & 30.1 & 0.826 & 46.7 & 0.912 & 0.368 & 0.850 \\ 
  Baseline-flat & 100.0 & 36.4 & 1.000 & 51.2 & 1.000 & 0.308 & 0.903 \\ 
   \hline
\\
\multicolumn{8}{l}{Experiment C: Data preprocessing} \\
  \hline
Model & \% Submitted & MWIS & RWIS & MAE & RAE & 50\% Cov. & 95\% Cov. \\ 
  \hline
GBQR-no-reporting-adj & 100.0 & \textbf{20.8} & \textbf{0.572} & \textbf{31.8} & \textbf{0.622} & 0.518 & 0.942 \\ 
  GBQR & 100.0 & 21.3 & 0.583 & 32.4 & 0.632 & 0.544 & \textbf{0.952} \\ 
  GBQR-no-transform & 100.0 & 22.1 & 0.606 & 34.0 & 0.664 & \textbf{0.496} & \textbf{0.948} \\ 
  Baseline-flat & 100.0 & 36.4 & 1.000 & 51.2 & 1.000 & 0.308 & 0.903 \\ 
   \hline
\end{tabular}
\caption{Evaluation results for post hoc experiments investigating determinants of model performance, omitting forecasts made on combinations of reference date and location for which the latest available NHSN data at the time of the forecast were subsequently revised by 10 or more admissions.  Experiment A gives results for individual component models in the Flusion ensemble, ensembles of pairs of components, and the full Flusion ensemble including all three components.  Experiment B gives results for the GBQR model, which is trained jointly on data for all locations and data sources, and variations trained separately for each location (GBQR-by-location) and trained only on hospital admissions from NHSN (GBQR-only-NHSN).  Experiment C gives results for a variation on the GBQR model that does not incorporate reporting adjustments designed to improve the degree to which ILINet and FluSurvNET data reflect influenza activity (GBQR-no-reporting-adj) and a variation that does not use a fourth-root transform (GBQR-no-transform), along with the original GBQR model which uses the reporting adjustments and the fourth-root transform. The percent of all combinations of location, reference date, and horizon for which the given model submitted forecasts is shown in the ``\% Submitted" column; in these retrospective experiments, we produced forecasts for all locations and time points. Within each experiment group, results for the model with the best MWIS, RWIS, MAE, and RAE are highlighted. Results for the models where empirical PI coverage rates are closest to the nominal levels are highlighted.} 
\label{tab:scores_experiments}
\end{table}

\printbibliography